\documentclass[10pt,twocolumn,letterpaper]{article}

\usepackage[pagenumbers]{cvpr}

\newcommand{\TODO}[1]{\textbf{\color{red}[TODO: #1]}}
\renewcommand{\TODO}[1]{}

\newcommand{\hlrow}{\rowcolor{black!6}}
\newcommand{\ours}{AnthroTAP\xspace}
\usepackage{colortbl}

\usepackage{multirow}

\usepackage{amsmath, amssymb, amsfonts} %
\usepackage{graphicx} %
\usepackage{bm} %
\usepackage{bbm}

\usepackage{caption}    %

\newcommand{\vect}[1]{\mathbf{#1}} %
\definecolor{dark_green}{rgb}{0, 0.5, 0}
\definecolor{light_purple}{rgb}{0.6, 0.5, 0.8}

\usepackage{xcolor}
\newcommand{\paragrapht}[1]{\noindent\textbf{#1}}

\definecolor{cvprblue}{rgb}{0.21,0.49,0.74}
\usepackage[pagebackref,breaklinks,colorlinks,allcolors=cvprblue]{hyperref}

\def\paperID{14390} %
\def\confName{CVPR}
\def\confYear{2026}

\title{AnthroTAP: Learning Point Tracking with Real-World Motion}

\usepackage[utf8]{inputenc} %
\usepackage[T1]{fontenc}    %
\usepackage{url}            %
\usepackage{booktabs}       %
\usepackage{amsfonts}       %
\usepackage{nicefrac}       %
\usepackage{microtype}      %
\usepackage{xcolor}         %

\usepackage{graphicx}
\usepackage{xspace}
\usepackage{colortbl}
\usepackage[dvipsnames]{xcolor}
\usepackage{pifont}

\author{
Inès Hyeonsu Kim$^{1,3}$\thanks{Equal contribution.} \quad Seokju Cho$^{1}$\footnotemark[1] \quad Jahyeok Koo$^{1}$ \quad Junghyun Park$^{1}$ \quad Jiahui Huang$^{2}$ \\ Honglak Lee$^{3,4}$ \quad Joon-Young Lee$^{2}$ \quad Seungryong Kim$^{1}$\\[0.5em]
$^{1}$KAIST AI \quad $^{2}$Adobe Research \quad $^{3}$University of Michigan \quad $^{4}$LG AI Research\\[1em]
\href{https://cvlab-kaist.github.io/AnthroTAP/}{https://cvlab-kaist.github.io/AnthroTAP/}
}

\begin{document}
\maketitle
\begin{abstract}
Point tracking models often struggle to generalize to real-world videos because large-scale training data is predominantly synthetic---the only source currently feasible to produce at scale. Collecting real-world annotations, however, is prohibitively expensive, as it requires tracking hundreds of points across frames. We introduce \textbf{AnthroTAP}, an automated pipeline that generates large-scale pseudo-labeled point tracking data from real human motion videos. Leveraging the structured complexity of human movement---non-rigid deformations, articulated motion, and frequent occlusions---AnthroTAP fits Skinned Multi-Person Linear (SMPL) models to detected humans, projects mesh vertices onto image planes, resolves occlusions via ray-casting, and filters unreliable tracks using optical flow consistency. A model trained on the AnthroTAP dataset achieves state-of-the-art performance on TAP-Vid, a challenging general-domain benchmark for tracking any point on diverse rigid and non-rigid objects (e.g., humans, animals, robots, and vehicles). 
Our approach outperforms recent self-training methods trained on vastly larger real datasets, while requiring only one day of training on 4 GPUs. 
AnthroTAP shows that structured human motion offers a scalable and effective source of real-world supervision for point tracking.

\end{abstract}

\section{Introduction}

\begin{figure}[t]
    \centering
    \includegraphics[width=\linewidth]{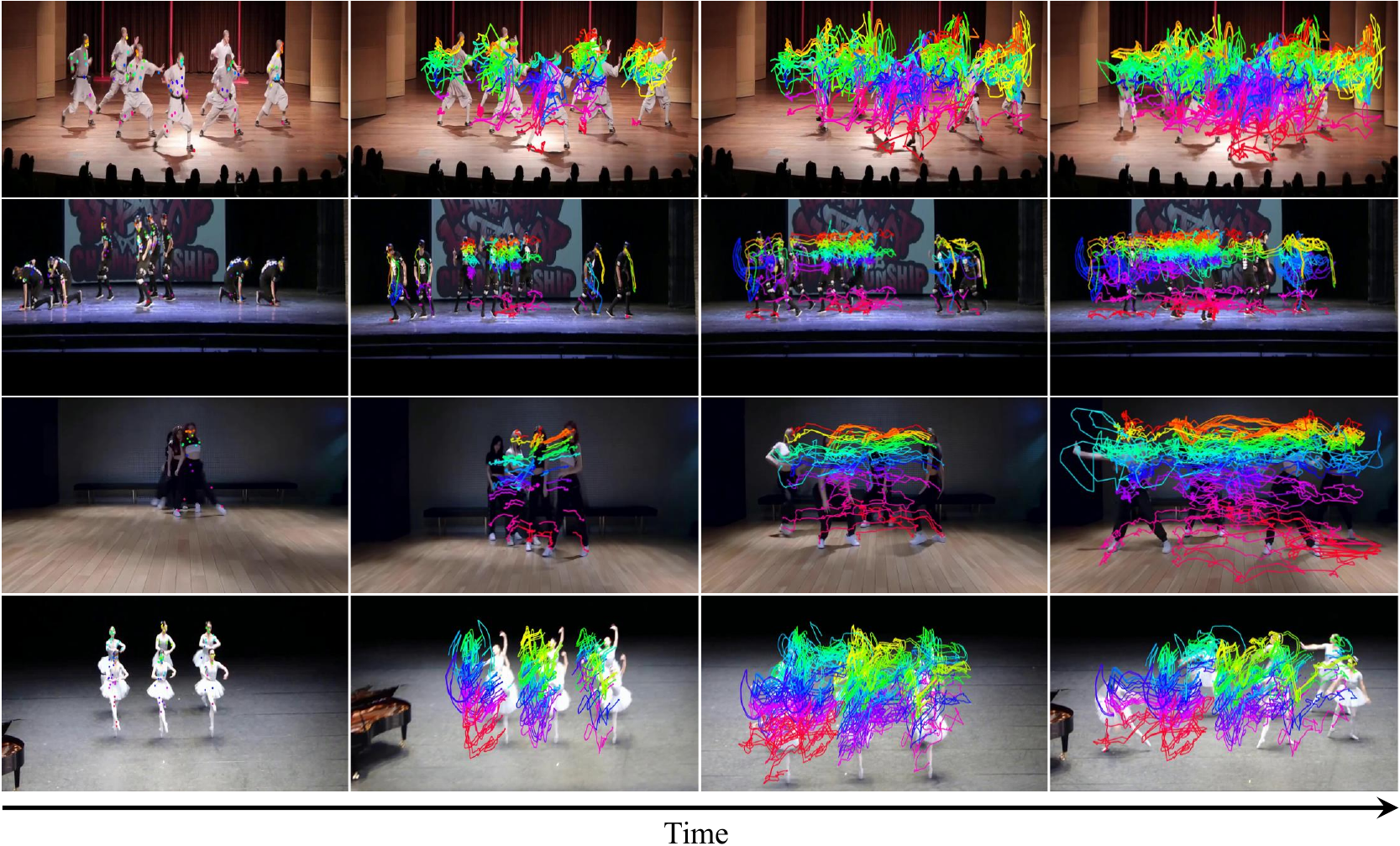}
    \vspace{-2em}
    \caption{\textbf{Visualization of videos annotated with our pipeline.} In this example, we visualize trajectories extracted using our pipeline using videos from~\cite{peize2021dance}. These examples highlight the complexity of human crowd movement, including both intra-person and inter-person interactions. We leverage this complexity as a source of supervision for training point tracking models.}
    \label{fig:video-visualization}
    \vspace{-1em}
\end{figure}

Accurately tracking points on an object's surface within a video, a task known as point tracking~\cite{doersch2022tap,harley2022particle,karaev2024cotracker, cho2024local,doersch2023tapir,karaev2024cotracker3,doersch2024bootstap,shrivastava2024gmrw}, has become important for various applications, including robotics~\cite{bharadhwaj2024track2act,vecerik2024robotap,yang2025magma,wen2024anypointtrajectorymodelingpolicy}, visual odometry~\cite{chen2024leap,rockwell2025dynpose}, 3D~\cite{wang2025vggt,wang2024vggsfm} or 4D reconstruction~\cite{wang2024shape,zhang2025tapip3d,seidenschwarz2024dynomo,badki2025l4p,kastenfast,st4rtrack2025}, video editing~\cite{geng2024motion,jeong2024track4gen}, and motion segmentation~\cite{huang2025segmentmotionvideos,karazija24learning}. 

Despite its growing significance, acquiring the extensive and diverse real-world data remains a significant challenge. Specifically, manually annotating point trajectories in video sequences is extremely labor-intensive and time-consuming~\cite{doersch2022tap,zheng2023pointodyssey}, making it a major bottleneck for acquiring real training data. 
Consequently, the limited availability of real-world training data often restricts the ability of current tracking models to generalize effectively across diverse real-world scenarios~\cite{karaev2024cotracker3,doersch2024bootstap}. Although synthetic datasets~\cite{mayer2016large,greff2022kubric,zheng2023pointodyssey} can be generated at scale, they often fail to capture the complex visual characteristics inherent to real-world scenarios.

Several recent works attempt to address this issue by utilizing self-training on real videos~\cite{doersch2024bootstap,karaev2024cotracker3}; however, these methods often demand extensive computational resources and vast datasets (e.g., 15M videos with 256 GPUs in Doersch et al.~\cite{doersch2024bootstap}) and often show limited improvement, possibly due to the weak supervision signal derived from the model itself. In addition, self-training frameworks may suffer from confirmation bias~\cite{sohn2020fixmatch}. As a result, efficiently training point tracking models with realistic video data remains an open challenge.

Our key insight is that human motion provides a strong supervision signal for point tracking. By distilling knowledge from human mesh fitting models, we can automatically establish reliable point correspondences in real videos and generate pseudo-labeled training data that captures non-rigid deformations, articulated movements, and frequent occlusions without manual annotation.
For instance, as an individual walks, dances, or turns, such a point can be obscured by limbs, or hidden by other articulated bodies in a crowded scene. When multiple people are present, the complexity increases further, as exemplified in Figure~\ref{fig:video-visualization}. In addition, videos capturing human activity reflect this real-world richness, featuring motion blur, varying lighting conditions, and reflective surfaces, all of which are highly difficult to replicate authentically in simulated environments. We believe that these challenges from human motion with real-world visual characteristics provide essential training signals for robust trackers. Furthermore, as shown in~\cite{Xiao_2024_CVPR,cho2025seurat}, point motion carries an inherent 3D structure. Given that 2D trajectories are projections of such 3D motion, leveraging the underlying human 3D geometry provides a natural and effective way to improve 2D tracking accuracy.

To this end, we introduce \textbf{\ours}, a pseudo-labeling pipeline that distills knowledge from human motion to automatically generate real-world point tracking training data. \ours initially fits the SMPL model to humans detected in each video frame using a pre-trained Human Mesh Recovery (HMR) method~\cite{dwivedi2024tokenhmr}, producing a 3D mesh for each person. The trajectories of these 3D mesh vertices are projected onto the 2D image plane, forming initial pseudo-trajectories. To manage occlusions, we apply ray-casting, leveraging the 3D mesh structure to estimate point visibility, considering both self-occlusion and inter-person occlusion. Finally, to enhance pseudo-label reliability, we introduce a filtering stage based on optical flow consistency~\cite{meister2018unflow} between adjacent frames. 

We validate our approach by training a point tracking model~\cite{cho2024local} using the generated pseudo-labels. The effectiveness of our data generation pipeline is demonstrated by the model's performance in complex real-world scenarios, achieving state-of-the-art results on the TAP-Vid benchmark~\cite{doersch2022tap}, which is a standard evaluation for tracking any point on diverse, real-world videos of general objects. Our method outperforms CoTracker3~\cite{karaev2024cotracker3}, which utilizes 11\(\times\) more training data, and BootsTAPIR~\cite{doersch2024bootstap}, which uses approximately 10,000\(\times\) more videos. This underscores the advantage of leveraging complex human motion for efficient and effective point tracking. Moreover, unlike previous works such as BootsTAPIR~\cite{doersch2024bootstap} and CoTracker3~\cite{karaev2024cotracker3} that uses proprietary videos, our dataset is non-proprietary, contributing valuable resources to the research community.

Our contributions are summarized as follows: 

\begin{itemize}
\item Introduces a simple yet effective approach to leverage real-world training data for point tracking by distilling knowledge from human motion, enabling automatic generation of pseudo-labeled datasets with strong supervision signals that capture complex motion patterns and real-world visual characteristics.
\item Leverages the inherent complexities of human motion from videos, using human model fitting and 3D geometry, to automatically generate pseudo-labeled datasets with complex, diverse trajectories, quantified by proposed trajectory complexity and diversity metrics.
\item Achieves state-of-the-art point tracking performance with significantly less training data, utilizing a non-proprietary dataset to offer a valuable contribution to the point tracking community.
\end{itemize}

\begin{figure*}[t]
    \centering
    \includegraphics[width=\textwidth]{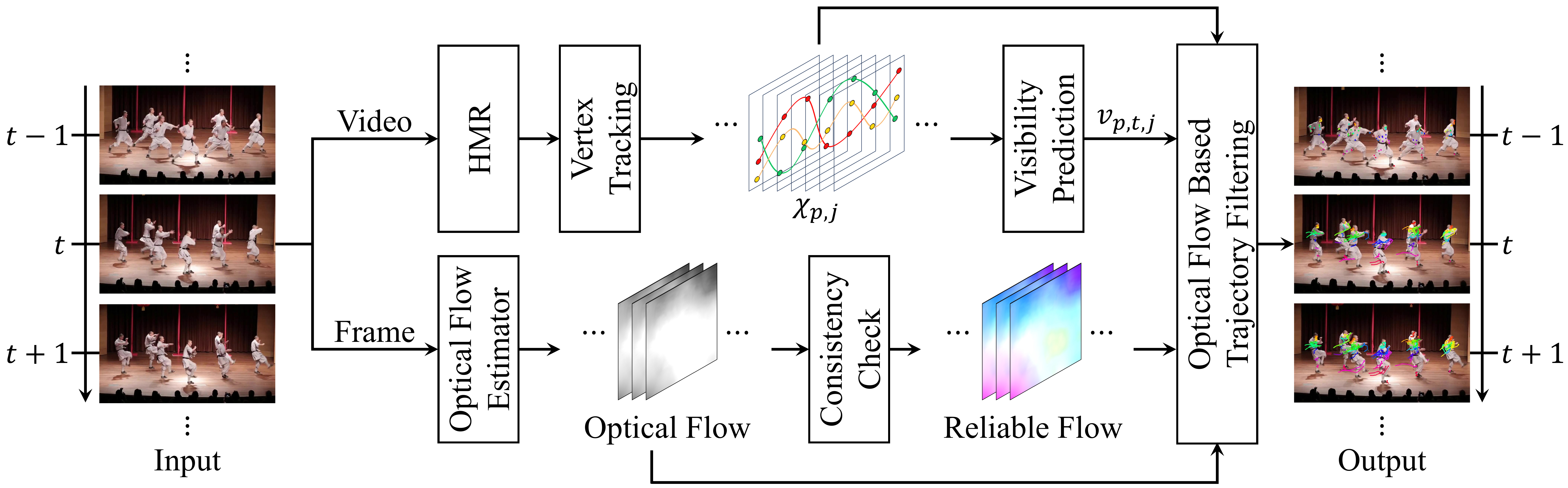}
    \vspace{-1em}
    \caption{\textbf{Overall pipeline.} We extract human meshes using an off-the-shelf human mesh recovery model~\cite{dwivedi2024tokenhmr}, and track points by projecting the mesh vertices. We also determine point visibility using ray-casting. In parallel, we extract optical flow~\cite{wang2024sea} and retain only reliable flow using forward-backward consistency~\cite{meister2018unflow}. Finally, we filter the trajectories by checking the consistency between the optical flow and the trajectories generated from the human mesh.}
    \label{fig:main_arch}
    \vspace{-1em}
\end{figure*}

\section{Related work}

\paragraph{Point tracking.}
PIPs~\cite{harley2022particle} iteratively refines point estimations by constructing local correlation maps, inspired by optical flow methods such as RAFT~\cite{teed2020raft}. TAPIR~\cite{doersch2023tapir} extends this idea by combining per-frame initialization from TAP-Net~\cite{doersch2022tap} with iterative refinement inspired by PIPs. CoTracker~\cite{karaev2024cotracker} learns to track multiple points jointly to enhance reliability. LocoTrack~\cite{cho2024local} introduces a local 4D correlation to establish better correspondence within videos.

Despite recent advances in model architectures for track refinement, existing methods~\cite{harley2022particle,doersch2022tap,doersch2023tapir,cho2024local,li2024taptr,li2024taptrv2,qu2024taptrv3,cho2024flowtrack,lemoing2024dense,aydemir2025track} predominantly rely on synthetic data~\cite{mayer2016large,greff2022kubric,zheng2023pointodyssey} during training, which can limit the robustness of point tracking in complex real-world scenarios. To alleviate this issue, Chrono~\cite{kim2025exploring} leverages representations learned by DINOv2~\cite{oquab2023dinov2}, a model pre-trained on large-scale real-world datasets. Similarly, BootsTAP~\cite{doersch2024bootstap} and CoTracker3~\cite{karaev2024cotracker3} incorporate unlabeled real-world data via self-training to further reduce dependency on synthetic data. However, these methods do not explicitly model the distinctive complexities found in human motion, such as significant non-rigid deformations, diverse textures from clothes, and frequent occlusions arising from interactions within and between individuals. In contrast, our approach specifically targets learning from human motion, which includes these challenging yet commonly occurring real-world phenomena.

\paragrapht{Point tracking datasets and benchmarks.}
Kubric~\cite{greff2022kubric} is a data generator that creates synthetic scenes for various tasks~\cite{sajjadi2022scene, singh2022simple, ren2025gen3c}, including point tracking. While most point tracking methods~\cite{doersch2023tapir, karaev2024cotracker, li2024taptr, cho2024local, lemoing2024dense} are trained solely on the Kubric dataset, it features limited motion, primarily consisting of objects falling onto the floor.
TAP-Vid~\cite{doersch2022tap} provides manual annotations on unlabeled real-world datasets~\cite{kay2017kinetics, lee2021beyond, pont20172017}. However, manually annotating point tracks in real videos is extremely time-consuming, making it impractical to scale such efforts for generating training data. DriveTrack~\cite{balasingam2024drivetrack} explores automated dataset generation in driving scenes~\cite{Sun_2020_CVPR}; however, the resulting trajectories lack complexity due to relatively simple vehicle motions, with no deformation or articulated motion.

\paragrapht{Human Mesh Recovery.}
The goal of Human Mesh Recovery (HMR) is to estimate a full 3D human mesh from a monocular 2D image or video. SMPL~\cite{loper2023smpl} proposes an optimization method for parameterized human pose and shape. Early works~\cite{bogo2016keep, lassner2017unite} leverage 2D joint prediction obtained from off-the-shelf 2D pose estimation models~\cite{cao2017realtime, xu2022vitpose} and predict 3D body model defined by SMPL. With the advancement of 3D body model prediction from single images, several video-based approaches~\cite{dwivedi2024tokenhmr, goel2023humans} have been developed. TokenHMR~\cite{dwivedi2024tokenhmr} introduces tokenized representation for human pose and achieves state-of-the-art accuracy on multiple in-the-wild 3D benchmarks. We aim to utilize the 3D prior from the Human Mesh Recovery model to automatically generate complex pseudo-labeled datasets and improve point trackers with the annotated datasets. 

\section{Method}

The development of robust point tracking models is often limited by the scarcity of large-scale, diverse real-world training data. This scarcity stems from the prohibitively time-consuming and labor-intensive nature of manually annotating point trajectories in real-world videos~\cite{doersch2022tap}, making large-scale annotation infeasible. Although synthetic datasets~\cite{mayer2016large,greff2022kubric,zheng2023pointodyssey,karaev2023dynamicstereo} can be generated at scale, they often fail to capture key characteristics of real-world scenes, such as complex {motions and} lighting, realistic appearance variations, and dynamic occlusions. Recent efforts have attempted to overcome this limitation using self-training approaches on unlabeled real videos~\cite{karaev2024cotracker3,doersch2024bootstap}. While promising, these methods suffer from weak supervision signals and often rely on random or heuristic trajectory sampling, limiting their training efficiency.

To address this challenge, we propose \textbf{\ours}, a pseudo-labeling pipeline that enables leveraging human motion as a natural source of rich supervision for point tracking. Human activities inherently involve complex visual and geometric phenomena, such as non-rigid deformations, articulated motion, occlusions, and diverse clothing appearances, as shown in Figure~\ref{fig:video-visualization}. To benefit from these, we leverage point trajectories across time, using the Skinned Multi-Person Linear (SMPL) model~\cite{SMPL:2015}, a parametric 3D representation of the human body, to generate complex pseudo-labels. Since each vertex on the SMPL mesh corresponds to a fixed anatomical location, it enables consistent and temporally stable point trajectories to be generated from human-centric videos.

To further improve the reliability of the resulting pseudo-labels, we introduce a refinement step based on short-range optical flow consistency, a method proven effective over short temporal windows~\cite{cho2024flowtrack}. This process filters out erroneous trajectory segments caused by mesh fitting errors or occlusions from objects not modeled by SMPL, such as furniture or scene clutter.

An overview of our full pipeline is shown in Figure~\ref{fig:main_arch}, and the following sections describe each stage in detail, from initial pseudo-label generation to final trajectory refinement.

\subsection{Point tracking with human mesh models}
Our pseudo-label generation process is founded upon utilizing the SMPL model~\cite{SMPL:2015} as a representation for human bodies. For each detected person, indexed by $p$, in a video frame $t$, the SMPL model defines a 3D body mesh $\mathcal{M}_{p,t}$ composed of $N_v$ vertices. This mesh is parameterized by low-dimensional 3D human shape parameters $\bm{\beta}_{p} \in \mathbb{R}^{D_{\beta}}$ (typically assumed constant for a person across a sequence) and 3D human pose parameters $\bm{\theta}_{p,t} \in \mathbb{R}^{D_{\theta}}$. $D_{\beta}$ represents the dimensionality of the identity-specific shape space, while $D_{\theta}$ corresponds to the dimensionality of the pose space, capturing body articulation and global orientation. The function $M(\bm{\beta}_{p}, \bm{\theta}_{p,t})$ maps these parameters to a set of 3D vertex locations $\{\vect{v}_{p,t,j} \in \mathbb{R}^3\}_{j=1}^{N_v}$.

To obtain these SMPL parameters from video frames, we employ a pre-trained video Human Mesh Recovery (HMR) method~\cite{dwivedi2024tokenhmr}. Leveraging a strong human prior, the model can reliably track and reconstruct human bodies even under motion blur, extreme motion, or occlusions, which are scenarios where conventional point tracking methods often fail. Given an input video $\{I_t\}_{t=1}^{T}$, where $T$ is the number of frames, the HMR model processes the video and outputs the estimated shape parameters $\bm{\beta}_{p}$ and pose parameters $\{\bm{\theta}_{p,t}\}_{t=1}^{T}$ for each detected person $p$. 
When the HMR model is reliable across a sequence of frames, the 3D positions of a fixed vertex $j$ on person $p$, denoted by $\{\vect{v}_{p,t,j}\}_{t \in \mathcal{T}_{p,j}}$, define an initial 3D trajectory of a point on the human body surface.
Here, $\mathcal{T}_{p,j}$ represents the set of frame indices for which vertex $j$ of person $p$ is successfully tracked.

These estimated 3D vertex locations are subsequently projected onto the 2D image plane to generate corresponding 2D trajectories. Let $\Pi: \mathbb{R}^3 \rightarrow \mathbb{R}^2$ denote the camera projection function, which maps a 3D point in camera coordinates to its 2D pixel coordinates using known camera intrinsic and extrinsic parameters. The 2D pseudo-trajectory for vertex $j$ of person $p$ is thus given by $\mathcal{X}_{p,j} = \{\vect{x}_{p,t,j} = \Pi(\vect{v}_{p,t,j}) \mid t \in \mathcal{T}_{p,j}\}$.

\paragrapht{Visibility prediction with ray casting.}
Accurate prediction of point visibility is important for generating reliable pseudo-labels, as points that become occluded, either by parts of the same body or by other humans or scene elements, cannot provide dependable supervision signals. Given explicit 3D mesh representations for all detected humans, we determine the estimated visibility $v_{p,t,j}$ for each 2D pseudo-labeled point $\vect{x}_{p,t,j}$ based on whether a ray from the camera center to its corresponding 3D vertex $\vect{v}_{p,t,j}$ intersects any triangle from any human mesh $\mathcal{M}_{p',t}$ before reaching the target vertex. If no such intersection occurs, we mark the point as visible ($v_{p,t,j}=1$); otherwise, we mark it as occluded ($v_{p,t,j}=0$). We implement this ray-triangle intersection test using the Möller-Trumbore algorithm~\cite{moller2005fast}, which runs efficiently on modern GPUs. This step handles both self-occlusion and occlusion by other people, but it does not capture occlusions from scene elements that are not represented by the human meshes, which we address in the subsequent filtering stage.

\subsection{Supervised training with pseudo-labels}
Vertex projection and ray casting for visibility prediction yield a collection of refined 2D point trajectories $\mathcal{X}_{p,j}$ with associated visibility labels $v_{p,j}$. These trajectories serve as additional supervision for existing point tracking models~\cite{cho2024local,karaev2024cotracker3}, and we simply apply each model's original training loss. 

To further improve the quality of our pseudo-labeled human tracks, we introduce an additional optical flow-based filtering stage before training that identifies and removes unreliable trajectory segments, including those caused by occlusions from scene elements that are not represented by the human meshes.

\paragrapht{Optical-flow based trajectory filtering.} While HMR methods are generally robust, they do not model occlusions from scene objects such as furniture or background clutter; the predicted trajectory can therefore continue to follow the body surface even after the point has been occluded by such elements. Optical flow~\cite{teed2020raft}, by contrast, naturally reflects the true image motion: when a point is occluded, the flow either becomes inconsistent or tracks the occluder rather than the underlying surface, producing a detectable divergence from the HMR prediction. We exploit this discrepancy by comparing the HMR-predicted displacement at each tracked point with the corresponding flow-derived displacement.

We first apply a forward-backward consistency check~\cite{meister2018unflow} between each pair of consecutive frames to identify transitions with locally reliable flow. For each trajectory, we then flag transitions where the HMR-predicted displacement and the optical flow displacement diverge beyond a threshold, after normalizing both vectors by the shorter of the two magnitudes. This naturally captures occlusions by unmodeled scene elements: when a point is occluded, the local flow either fails the consistency check or follows the occluder's motion rather than the underlying body surface, causing a mismatch with the HMR prediction. We then compute the fraction of flagged transitions per trajectory, restricted to frames where the point is marked visible by ray casting and the flow is reliable; trajectories exceeding a predefined error ratio are discarded entirely, while only the inconsistent transitions are removed from the remainder. The result is a smaller but substantially cleaner set of pseudo-labels. Full details appear in Sec.~C of the supplementary material.

\begin{table*}[t]
  \centering
  \caption{\textbf{Quantitative results on TAP-Vid benchmark~\cite{doersch2022tap} and RoboTAP~\cite{vecerik2024robotap}.} Both LocoTrack and TAPNext, trained with our approach (Anthro-LocoTrack and Anthro-TAPNext), show significant performance improvements over their respective baselines across all metrics and datasets. We use a training dataset that is $11\times$ smaller than CoTracker3~\cite{karaev2024cotracker3} in terms of the number of videos, and $1{,}000\times$ smaller than the datasets used in BootsTAPIR~\cite{doersch2024bootstap} in terms of the number of training frames. Kub refers to the Kubric dataset~\cite{greff2022kubric}, while Kub64 refers to the dataset rendered in Karaev et al.~\cite{karaev2024cotracker3}. Both Kub and Kub64 are synthetic datasets.}
  \label{tab:tapvid_results}
  \resizebox{\textwidth}{!}{
    \begin{tabular}{ll|ccc|ccc|ccc|ccc}
    \toprule
    \multirow{2}{*}{Method}& Training& \multicolumn{3}{c|}{\textbf{DAVIS First}} & \multicolumn{3}{c|}{\textbf{DAVIS Strided}} & \multicolumn{3}{c|}{\textbf{Kinetics First}} & \multicolumn{3}{c}{\textbf{RoboTAP First~\cite{vecerik2024robotap}}}\\
     & Dataset & AJ & $<\delta_{avg}^x$  & OA  & AJ & $<\delta_{avg}^x$  & OA  & AJ & $<\delta_{avg}^x$  & OA  & AJ & $<\delta_{avg}^x$  & OA  \\
     \midrule
     \multicolumn{14}{c}{Models evaluated at $256\times 256$ resolution} \\

    OmniMotion \citep{wang2023tracking} & - & - & - & - & $51.7$ & $67.5$ & $85.3$ & - & - & - & - & - & -\\
    Dino-Tracker \citep{tumanyan2024dino} & - & -  & -  & - & $62.3$ & $78.2$ & $87.5$ & - & - & - & - & - & - \\
    \midrule
    TAPNet \citep{doersch2022tap} & Kub & $33.0$ & $48.6$ & $78.8$ & $38.4$ & $53.1$ & $82.3$ & $38.5$ & $54.4$ & $80.6$ & - & - & - \\
    TAPIR \citep{doersch2023tapir} & Kub & $58.5$ & $70.0$ & $86.5$ & $61.3$ & $73.6$ & $88.8$ & $49.6$ & $64.2$ & $85.0$ & $59.6$              & $73.4$                & $87.0$  \\
    Online TAPIR \citep{vecerik2024robotap} & Kub & $56.2$ & $70.0$ & $86.5$ & - & - & - & $51.5$ & - & - & - & - & - \\
    TAPTR \citep{li2024taptr} & Kub & $63.0$ & $76.1$ & $91.1$ & $66.3$ & $79.2$ & $91.0$ & $49.0$ & $64.4$ & $85.2$ & $60.1$              & $75.3$                & $86.9$ \\
    TAPTRv2 \citep{li2024taptrv2} & Kub & $63.5$ & $75.9$ & $\underline{91.4}$ & $66.4$ & $78.8$ & $91.3$ & $49.7$ & $64.2$ & $85.7$ & - & - & -\\
    TAPTRv3 \citep{qu2024taptrv3} & Kub & $63.2$ & $76.7$ & $91.0$ & - & - & - & $\underline{54.5}$ & $\underline{67.5}$ & $\textbf{88.2}$ & - & - & -\\
    BootsTAPIR \citep{doersch2024bootstap}& Kub+15M & $61.4$ & $74.0$ & $88.4$ & $66.2$ & $78.5$ & $90.7$ & $\textbf{54.6}$ & $\textbf{68.4}$ & $\underline{86.5}$ & $\textbf{64.9}$  & $\textbf{80.1}$       & $86.3$ \\
    \midrule
    LocoTrack \cite{cho2024local} & Kub & $63.0$ & $75.3$ & $87.2$ & $67.8$ & $79.6$ & $89.9$ & $52.9$ & $66.8$ & $85.3$ & $62.3$              & $76.2$                & $87.1$ \\
    \multirow{1}{*}{Anthro-LocoTrack (\textbf{Ours})} & \multirow{1}{*}{Kub+1.4K}& $64.8$ & $77.3$ & $89.1$ & $\underline{69.0}$ & $81.0$ & $90.8$ & $53.9$ & $\textbf{68.4}$ & $86.4$ & $\underline{64.7}$              & $\underline{79.2}$                & $88.4$\\

    \hlrow\multicolumn{2}{c|}{Improvement over baseline} & \textcolor{ForestGreen}{+1.8} & \textcolor{ForestGreen}{+2.0} & \textcolor{ForestGreen}{+1.9} & \textcolor{ForestGreen}{+1.2} & \textcolor{ForestGreen}{+1.4} & \textcolor{ForestGreen}{+0.9} & \textcolor{ForestGreen}{+1.0} & \textcolor{ForestGreen}{+1.6} & \textcolor{ForestGreen}{+1.1} & \textcolor{ForestGreen}{+2.4} & \textcolor{ForestGreen}{+3.0} & \textcolor{ForestGreen}{+1.3}\\
    \midrule
    TAPNext \cite{zholus2025tapnext} & Kub & $62.4$ & $76.6$ & $90.5$ & $65.4$ & $79.7$ & $88.9$ & - & - & - & 59.8 & 73.1 & 88.1 \\
    BootsTAPNext \cite{zholus2025tapnext} & Kub+15M & $\underline{65.2}$ & $\underline{78.5}$ & $91.2$ & $68.9$ & $\underline{82.4}$ & $\underline{91.6}$ & - & - & - & 64.1 & 75.1 & $\underline{88.8}$ \\
    \multirow{1}{*}{Anthro-TAPNext (\textbf{Ours})} & \multirow{1}{*}{Kub+1.4K} & $\textbf{66.1}$ & $\textbf{79.3}$ & $\textbf{91.7}$ & $\textbf{71.4}$ & $\textbf{83.5}$ & $\textbf{92.4}$ & - & - & - & 63.4 & 76.3 & $\textbf{90.2}$ \\
    \hlrow\multicolumn{2}{c|}{Improvement over baseline} & \textcolor{ForestGreen}{+3.7} & \textcolor{ForestGreen}{+2.7} & \textcolor{ForestGreen}{+1.2} & \textcolor{ForestGreen}{+6.0} & \textcolor{ForestGreen}{+3.8} & \textcolor{ForestGreen}{+3.5} & - & - & - & \textcolor{ForestGreen}{+3.6} & \textcolor{ForestGreen}{+3.2} & \textcolor{ForestGreen}{+2.1}\\

    \midrule\midrule
    \multicolumn{14}{c}{Models evaluated at $384\times 512$ resolution} \\
    PIPs~\citep{harley2022particle} & FT~\cite{mayer2016large} &  $42.2$ & $64.8$ & $77.7$ & $52.4$ & $70.0$ & $83.6$ & - & - & - & - & - & - \\
    CoTracker2~\citep{karaev2024cotracker} & Kub & $62.2$ & $75.7$ & $89.3$ & $65.9$ & $79.4$ & $\underline{89.9}$ & $48.8$ & $64.5$ & $85.8$ & - & - & - \\
    Track-On \citep{aydemir2025track} & Kub & $\underline{65.0}$ & $\underline{78.0}$ & $\underline{90.8}$ & - & - & - & $53.9$ & $67.3$ & $\underline{87.8}$ & - & - & - \\
    CoTracker3 (online)~\citep{karaev2024cotracker3} & Kub64+15K& $64.4$ & $76.9$ & $\textbf{91.2}$ & - & - & - & $54.7$ & $67.8$ & $87.4$ & - & - & - \\
    CoTracker3 (offline)~\citep{karaev2024cotracker3} & Kub64+15K & $63.8$ & $76.3$ & $90.2$ & - & - & - & $\textbf{55.8}$ & $\underline{68.5}$ & $\mathbf{88.3}$ & - & - & - \\
    \midrule
    LocoTrack~\cite{cho2024local} & Kub & $64.8$ & $77.4$ & $86.2$ & $\underline{69.4}$ & $\underline{81.3}$ & $88.6$ & $52.3$ & $66.4$ & $82.1$ & - & - & - \\
    \multirow{1}{*}{Anthro-LocoTrack (\textbf{Ours})} & \multirow{1}{*}{Kub+1.4K} & $\textbf{65.9}$ & $\textbf{78.9}$ & $87.3$ & $\mathbf{71.1}$ & $\textbf{82.9}$ & $\textbf{90.3}$ & $\underline{54.8}$ & $\textbf{68.6}$ & $85.3$ & - & - & -\\
    \hlrow\multicolumn{2}{c|}{Improvement over baseline} & \textcolor{ForestGreen}{+1.1} & \textcolor{ForestGreen}{+1.5} & \textcolor{ForestGreen}{+1.1} & \textcolor{ForestGreen}{+1.7} & \textcolor{ForestGreen}{+1.6} & \textcolor{ForestGreen}{+1.7} & \textcolor{ForestGreen}{+2.5} & \textcolor{ForestGreen}{+2.2} & \textcolor{ForestGreen}{+3.2} & - & - & -\\

    \bottomrule
    \end{tabular}
}
\end{table*}

\section{Experiments}

\begin{figure*}[t]
    \centering
    \includegraphics[width=\linewidth]{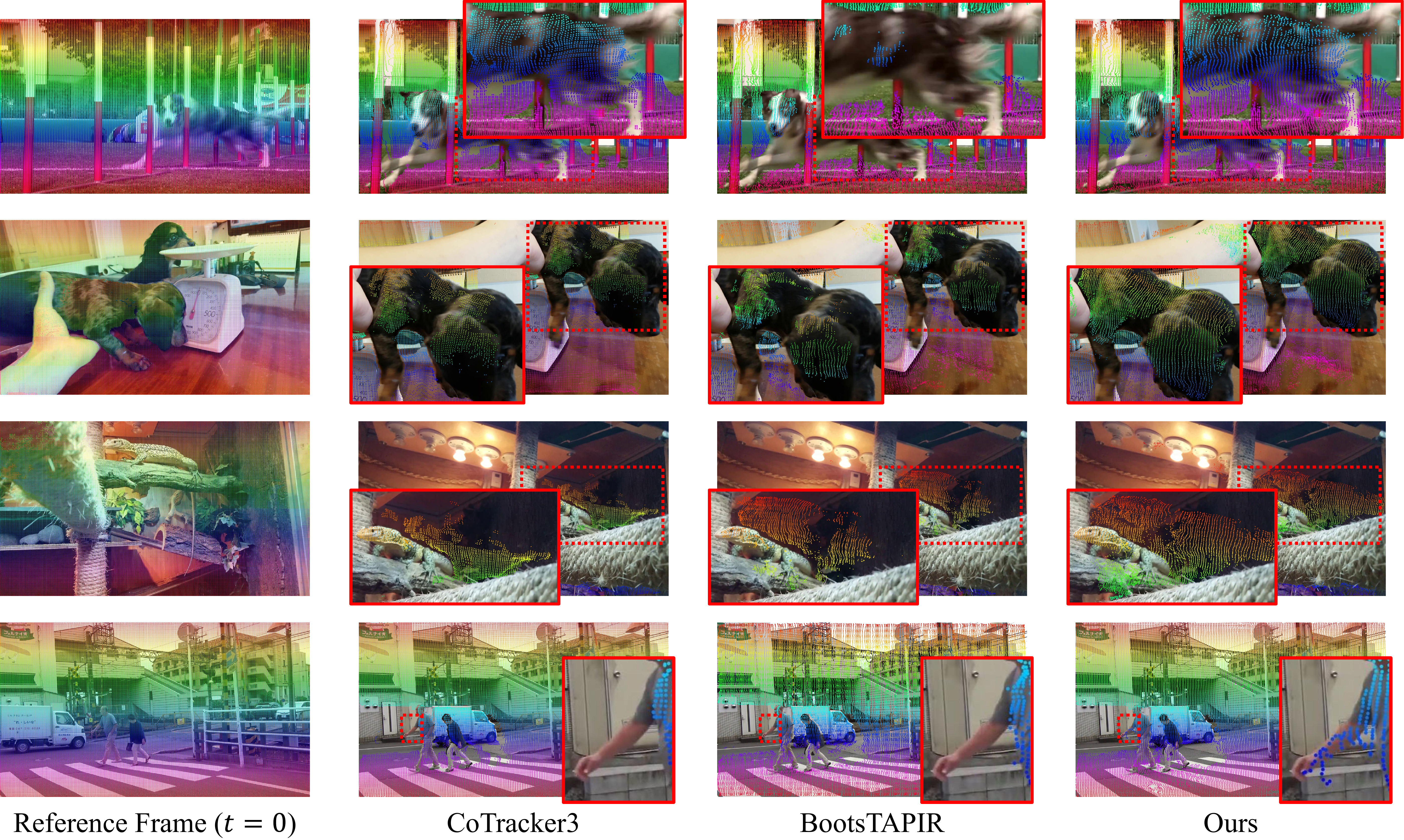}
    \vspace{-1em}
    \caption{\textbf{Qualitative results on the videos from DAVIS dataset~\cite{pont20172017}.} We present comparisons with CoTracker3~\cite{karaev2024cotracker3} and BootsTAPIR~\cite{doersch2024bootstap}. Anthro-LocoTrack (Ours) consistently demonstrates strong performance on highly deformable objects and severe occlusions.}
    \label{fig:qual_comparison}
\end{figure*}

\subsection{Experimental setup} 
\paragraph{Dataset construction.}
We use the pre-trained TokenHMR~\cite{dwivedi2024tokenhmr} model for human mesh recovery and SEA-RAFT~\cite{wang2024sea} for obtaining optical flow. To construct our training dataset, we use 1.4K videos from the Let's Dance dataset~\cite{castro2018let}, which includes a diverse range of dance performances, from solo acts to multi-person scenes. Unlike BootsTAP~\cite{doersch2024bootstap} or CoTracker3~\cite{karaev2024cotracker3}, our training videos are entirely non-proprietary, making our entire pipeline transparent and accessible.

\paragrapht{Training details.}
We fine-tune the pre-trained LocoTrack~\cite{cho2024local} base model using \ours-annotated Let's Dance dataset, which we call Anthro-LD. During fine-tuning, we randomly sample data from our dataset and Kubric Panning MOVi-E~\cite{doersch2023tapir} with equal probability. When training with our dataset, we do not supervise the occlusion prediction for points identified as occluded, as we intentionally filter the dataset for high precision. We augment the filtered dataset using affine transformations, following the similar approach used in BootsTAP~\cite{doersch2024bootstap}. Hyperparameters for the optical flow filtering are detailed in the supplementary material.

Optimization is performed using AdamW~\cite{loshchilov2017decoupled}, with the learning rate of $3 \times 10^{-4}$ and weight decay set to $1 \times 10^{-3}$. The model is trained for 50K steps. We employ a cosine learning rate schedule with a 1000-step warmup, and apply gradient clipping with a maximum norm of 1.0. Training is conducted with a batch size of 1 per GPU, using 4 NVIDIA A6000 GPUs, and converges within 1 day. This is significantly more efficient compared to CoTracker3~\cite{karaev2024cotracker3}, which uses 32 A100 80GB GPUs for training, and BootsTAPIR, which uses 256 A100 GPUs. For each batch, we randomly sample 256 tracks and use a resolution of $256 \times 256$.

\paragrapht{Evaluation protocol.}
We use the TAP-Vid~\cite{doersch2022tap} benchmark to evaluate our model trained on Anthro-LD. We also evaluate our model on the RoboTAP dataset~\cite{vecerik2024robotap}, a real-world robot dataset featuring videos of robotic manipulation tasks. The benchmark uses two evaluation modes based on how query points are selected: the \textbf{First} setup samples the query point from the first visible frame of each trajectory, whereas the \textbf{Strided} setup samples the query point every five timesteps along the trajectory. For evaluation, we use the following metrics, proposed in Doersch et al.~\cite{doersch2022tap}: Average Jaccard (AJ), position accuracy (\(<\delta_{avg}^x\)), and Occlusion Accuracy (OA).

\subsection{Main results}
\paragraph{Quantitative comparisons.} In Table~\ref{tab:tapvid_results}, we present quantitative results on the TAP-Vid benchmark applying our pipeline to two base models: LocoTrack~\cite{cho2024local} and TAPNext~\cite{zholus2025tapnext}. Our primary comparisons are against BootsTAPIR~\cite{doersch2024bootstap}, CoTracker3~\cite{karaev2024cotracker3}, and BootsTAPNext~\cite{zholus2025tapnext}. These baselines were trained on 15 million real videos (BootsTAPIR) and 15k real videos (CoTracker3), respectively, whereas our models are trained on just 1.4k videos.

Despite this significant difference in training data scale, our method achieves notable performance gains. On the DAVIS First, Anthro-LocoTrack surpasses BootsTAPIR by 3.3 percentage points on the $<\delta^x_{avg}$, demonstrating the data efficiency of our approach. Additionally, it outperforms CoTracker3 by 2.0 percentage points on the same metric. Applying our pipeline to TAPNext yields even larger improvements: Anthro-TAPNext surpasses BootsTAPNext by 2.5 percentage points on DAVIS Strided AJ.

\begin{table*}[t]
  \centering
  \captionof{table}{\textbf{Can training on human points generalize to non-human points?} To answer this question, we compare the performance by grouping query points on humans and non-human regions separately in TAP-Vid-DAVIS dataset~\cite{doersch2022tap}. Our method shows greater improvement on non-human points.}
  \label{tab:human-points}
\resizebox{0.75\textwidth}{!}{
      \begin{tabular}{l|l|ccc}
        \toprule
        Benchmark & Method & AJ &$<\delta^{x}_{avg}$ & OA \\
        \midrule\midrule
        \multirow{2}{*}{DAVIS (Human Only)} & LocoTrack & \underline{50.7} & \underline{42.4} & \underline{57.8} \\ %
        &  Anthro-LocoTrack (\textbf{Ours}) &  \textbf{51.2} (\textcolor{ForestGreen}{+0.5}) & \textbf{43.3} (\textcolor{ForestGreen}{+0.9}) & \textbf{58.6} (\textcolor{ForestGreen}{+0.8})  \\
        \midrule
        \multirow{2}{*}{DAVIS (Non-Human)} & LocoTrack & \underline{58.8} & \underline{73.1} & \underline{83.9} \\
        &  Anthro-LocoTrack (\textbf{Ours}) & \textbf{60.8} (\textcolor{ForestGreen}{+2.0}) & \textbf{75.2} (\textcolor{ForestGreen}{+2.1}) & \textbf{85.3} (\textcolor{ForestGreen}{+1.4}) \\
        \bottomrule
      \end{tabular}
    }
\end{table*}

\begin{table*}[t]
  \centering
  \captionof{table}{\textbf{Comparison with CoTracker3 self-training on identical training data.} We fine-tune CoTracker3 on the Let's Dance dataset~\cite{castro2018let} using two strategies: \textbf{(II)} the self-training strategy from CoTracker3~\cite{karaev2024cotracker3}, and \textbf{(III)} pseudo-labels generated by \ours. Our pipeline achieves substantially larger gains over the Kubric-pretrained baseline \textbf{(I)} than CoTracker3 self-training on the same data.}
  \label{tab:grid_capof_cotracker3_comparison}
\resizebox{0.75\textwidth}{!}{
      \begin{tabular}{l|lll|ccc}
        \toprule
        & \multirow{2}{*}{Model} & \multirow{2}{*}{Train Strategy} & \multirow{2}{*}{Dataset} & \multicolumn{3}{c}{\textbf{DAVIS}} \\
        &&& & AJ & $<\delta^{x}_{avg}$ & OA \\
        \midrule\midrule
        \textbf{(I)} & CoTracker3 & Supervised & Kubric & 63.8 & 76.3 & \underline{90.2}\\
        \textbf{(II)} & CoTracker3 & Self-training & LD~\cite{castro2018let} & \underline{64.2} (\textcolor{ForestGreen}{+0.4}) & \underline{76.5} (\textcolor{ForestGreen}{+0.2}) & 89.6 (\textcolor{BrickRed}{-0.6}) \\
        \midrule
        \hlrow \textbf{(III)} & CoTracker3 & \ours (\textbf{Ours}) & Anthro-LD & \textbf{65.0} (\textcolor{ForestGreen}{+1.2}) & \textbf{77.3} (\textcolor{ForestGreen}{+1.0}) & \textbf{90.7} (\textcolor{ForestGreen}{+0.5}) \\
        \bottomrule
      \end{tabular}
    }
\end{table*}

\paragrapht{Qualitative comparisons.} In Figure~\ref{fig:qual_comparison}, we visually compare the results of CoTracker3~\cite{karaev2024cotracker3}, BootsTAPIR~\cite{doersch2024bootstap}, and LocoTrack trained with our method. Our model demonstrates more robust tracking performance even in drastic deformation and severe occlusions. Refer to Sec.~B of the supplementary material for additional visualization.

\subsection{Ablation and analysis}

\paragraph{Can training on human points generalize to non-human points?}
In Table~\ref{tab:human-points}, we investigate this question by splitting the query points in the TAP-Vid-DAVIS dataset into human and non-human categories, and comparing the performance gain in each group relative to the baseline. We use Mask2Former~\cite{cheng2021mask2former} to identify points on humans. The results show that the performance improvements are significant for both human and non-human points, with an even greater boost observed for non-human points.

\paragrapht{Comparison with self-training strategy of CoTracker3 on identical data.}
Table~\ref{tab:grid_capof_cotracker3_comparison} compares our approach with the self-training strategy of CoTracker3~\cite{karaev2024cotracker3} on identical data. We fine-tuned the Kubric-pretrained CoTracker3 baseline on the Let's Dance dataset~\cite{castro2018let}, the same data our pipeline employs, via two methods: the original self-training from CoTracker3, and fine-tuning with pseudo-labels from \ours. This allowed a direct comparison of both data utilization strategies on the same baseline and data.

Results in Table~\ref{tab:grid_capof_cotracker3_comparison} show self-training by CoTracker3 yielded marginal baseline improvement on Let's Dance. In contrast, pseudo-labels from \ours achieved significantly better performance, indicating that our pseudo-labeling pipeline provides more effective supervision on this dataset than the CoTracker3 self-training strategy.

\paragrapht{Comparative analysis on trajectory complexity and diversity.} In Table~\ref{tab:new-metric}, we provide a comparative analysis of prevalent point tracking training datasets, including Kubric~\cite{greff2022kubric}, PointOdyssey~\cite{zheng2023pointodyssey}, and DriveTrack~\cite{balasingam2024drivetrack}, focusing on trajectory complexity and diversity. Complexity is measured using the mean angular acceleration, which reflects how sharply and frequently a trajectory changes direction over all contiguous visible segments. Diversity is assessed by the mean standard deviation of trajectory shapes within each video, independent of absolute position. Anthro-LD achieves the highest complexity by a large margin, surpassing even synthetic datasets such as PointOdyssey and Kubric that are designed with diverse object motion. DriveTrack, despite being a real-world dataset, shows substantially lower diversity, as its tracks are concentrated on rigid objects undergoing largely uniform motion. These results suggest that human body motion in dance videos offers richer and more varied trajectory patterns compared to other real-world datasets. Refer to Sec.~D of the supplementary material for details.

\begin{table}[t]
  \centering
  \captionof{table}{\textbf{Comparison of track complexity and diversity with existing datasets.} We compare trajectory complexity and diversity across real and synthetic point tracking datasets. Anthro-LD achieves the highest complexity among all datasets and substantially greater diversity than DriveTrack, the only other real-world dataset.}
  \resizebox{0.8\linewidth}{!}{%
      \begin{tabular}{l|l|cc}
        \toprule
        \multirow{2}{*}{Training Dataset} & \multirow{2}{*}{Data Type} & \multicolumn{2}{c}{\textbf{Traj Metrics}} \\
        & & Complexity & Diversity \\
        \midrule
        \midrule
        
        DriveTrack \citep{balasingam2024drivetrack}& Real  & {0.4396} & 0.0073   \\ %
        PointOdyssey \citep{zheng2023pointodyssey}& Synthetic & \underline{0.5222} & \textbf{0.1597}  \\ %
        Kubric \citep{greff2022kubric}& Synthetic & 0.1772 & \underline{0.1165} \\ %
 
        \midrule
        \hlrow Anthro-LD (\textbf{Ours}) & Real &  \textbf{1.2492} & {0.1008} \\ %
        \bottomrule
      \end{tabular}%
  }
  \label{tab:new-metric}
\end{table}

\begin{figure*}[t]
    \centering
    \includegraphics[width=\linewidth]{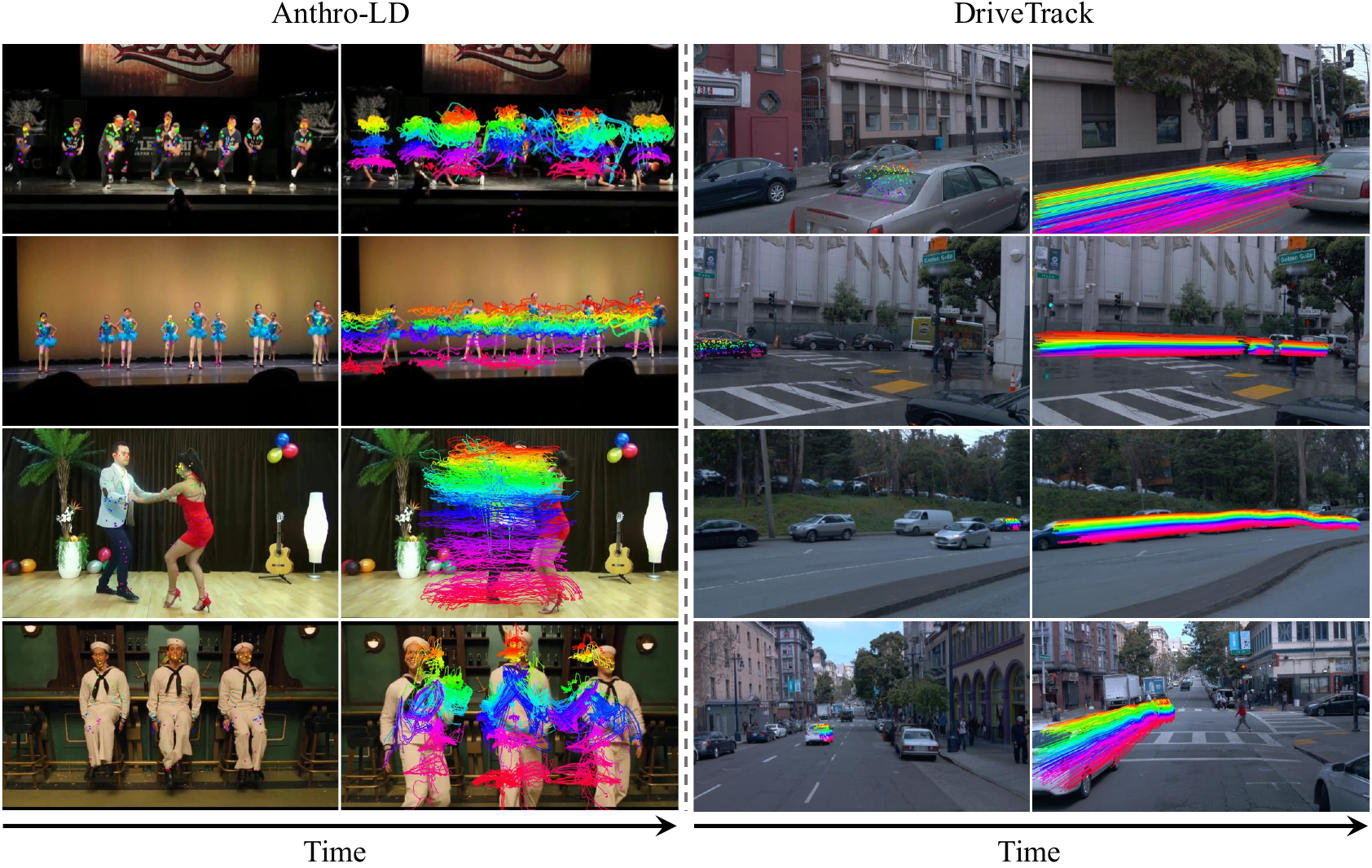}
    \vspace{-1em}
    \caption{\textbf{Qualitative comparison of trajectory complexity.} Anthro-LD, a dataset generated from our pipeline, exhibits diverse and complex motions, unlike the rigid and simple trajectories found in DriveTrack~\cite{balasingam2024drivetrack}.}
    \label{fig:data-qual-comparison}
    \vspace{-1em}
\end{figure*}

\paragrapht{Comparing our pseudo-labels with synthetic perfect GT.}
In Table~\ref{tab:blinkvision_locotrack}, we investigate whether annotation accuracy or real-world visual diversity is the more critical factor for improving point tracking. We compare against BlinkVision~\cite{li2024blinkvision}, a synthetic 3D dataset that includes human motion and provides perfect GT annotations. Training LocoTrack with BlinkVision yields moderate gains over the baseline, while our pseudo-labeled real-world data achieves substantially larger improvements. Given that both datasets contain human motion, the gap suggests that real-world appearance is a more decisive factor than annotation precision for learning generalizable tracking.

\begin{table}[t]
  \centering
  \caption{\textbf{Pseudo-GT on real data vs.\ perfect GT on synthetic data.} We compare LocoTrack fine-tuned on BlinkVision~\cite{li2024blinkvision} (perfect GT, synthetic) and our \ours-annotated Let's Dance dataset (pseudo-GT, real-world). Despite relying on pseudo-GT labels, our method achieves larger gains, indicating that real-world visual diversity matters more than annotation precision.}
  \resizebox{\linewidth}{!}{
  \begin{tabular}{l l | c c c}
    \toprule
    \multirow{2}{*}{Method} & \multirow{2}{*}{Datasets} & \multicolumn{3}{c}{\textbf{DAVIS}} \\
    & & AJ & $<\delta_{avg}$ & OA \\
    \midrule
    LocoTrack (256$\times$256) & Kub64 & 63.0 & 75.3 & 87.2 \\
    LocoTrack + BlinkVision~\cite{li2024blinkvision} & Kub64 + 1.4K & \underline{64.2} & \underline{76.4} & \underline{88.8} \\
    \hlrow LocoTrack + \textbf{Ours} & Kub64 + 1.4K & \textbf{64.8} & \textbf{77.3} & \textbf{89.1} \\
    \bottomrule
  \end{tabular}
}
  \label{tab:blinkvision_locotrack}
\end{table}

\paragrapht{Pseudo-label quality analysis.}
To assess the positional accuracy of our pseudo-labels, we collect human-annotated trajectories on the Let's Dance dataset as ground truth and evaluate position accuracy across five thresholds of 1, 2, 4, 8, and 16 pixels ($\delta^0$--$\delta^4$) following~\cite{doersch2022tap}, against both AnthroTAP-annotated and LocoTrack-predicted trajectories.
As shown in Table~\ref{tab:human_eval}, AnthroTAP-annotated trajectories substantially outperform LocoTrack-predicted ones, demonstrating that our pseudo-labels are of sufficient quality to serve as effective training supervision.

\begin{table}[t]
    \centering
    \captionof{table}{\textbf{Pseudo-label quality evaluated against human annotations.} We measure the positional accuracy of \ours pseudo-labels and LocoTrack predictions against human-annotated trajectories on the Let's Dance dataset. \ours pseudo-labels consistently outperform LocoTrack predictions across all thresholds, validating their quality as training supervision.}
    \label{tab:human_eval}
    \resizebox{\linewidth}{!}{
        \begin{tabular}{l|cccccc}
        \toprule
        \multirow{2}{*}{Method} & \multicolumn{6}{c}{\textbf{Human Annotated Trajs. on Let's Dance}}\\
        & $<\delta^{0}$ & $<\delta^{1}$ & $<\delta^{2}$ & $<\delta^{3}$ & $<\delta^{4}$ & $<\delta^{x}_{avg}$ \\
        \midrule\midrule
        LocoTrack & \underline{13.2} & \underline{36.2} & \underline{62.3} & \underline{81.0} & \underline{87.7} & \underline{56.1} \\
        \hlrow \textbf{AnthroTAP pseudo-label (Ours)} & \textbf{18.0} & \textbf{43.2} & \textbf{74.8} & \textbf{92.4} & \textbf{94.1} & \textbf{64.5} \\
        \bottomrule
        \end{tabular}
    }
    \vspace{-2em}
\end{table}

\paragrapht{Ablation on optical flow based track filtering.}
In Table~\ref{tab:grid_capof_flow_rejection}, we ablate the effect of optical flow-based filtering by comparing it with the baseline that uses trajectories projected from the human mesh and occlusion prediction via ray casting. While the baseline already achieves strong performance, applying trajectory rejection yields a further performance boost.

\paragrapht{Visual comparison with DriveTrack~\cite{balasingam2024drivetrack}.} In Figure~\ref{fig:data-qual-comparison}, we qualitatively compare our dataset with DriveTrack~\cite{balasingam2024drivetrack}, which is also a point-tracking dataset sourced from real videos. Observing the point traces, DriveTrack mostly exhibits monotonic motion, and the tracks tend to follow similar trajectories because they are concentrated on rigid objects undergoing largely uniform movement. In contrast, our dataset contains highly complex and diverse motion patterns.

\begin{table}
  \centering
  \captionof{table}{\textbf{Ablation on optical flow based track rejection.} The baseline uses trajectories projected from the human mesh with occlusion predicted via ray casting, without any filtering. Adding optical flow based rejection further improves performance across all metrics.}
  \resizebox{0.8\linewidth}{!}{%
      \begin{tabular}{l|ccc}
        \toprule
        \multirow{2}{*}{Method} & \multicolumn{3}{c}{\textbf{DAVIS}} \\
        & AJ &$<\delta^{x}_{avg}$ & OA \\
        \midrule\midrule
        Baseline & \underline{64.4} & \underline{76.9} & \underline{88.6}\\
        \hlrow + Optical flow based rejection  & \textbf{64.8} & \textbf{77.3} & \textbf{89.1}\\
        \bottomrule
      \end{tabular}%
  }
  \label{tab:grid_capof_flow_rejection}
  \vspace{-1em}
\end{table}

\vspace*{-0.1in}
\section{Conclusion}
In this paper, we presented a novel method for generating highly complex pseudo-labeled data for point tracking by leveraging the inherent complexities of human motion captured in videos. By fitting SMPL models to real-world human videos, accurately modeling occlusions via ray-casting, and refining trajectories using optical flow consistency, our approach significantly alleviates the bottleneck of manual annotation. A model trained on our dataset achieves state-of-the-art results with orders of magnitude less data than existing methods. Our method provides an effective path forward for robust point tracking in real-world scenarios.

\twocolumn[{%
    \begin{center}
        {\Large\textbf{AnthroTAP: Learning Point Tracking with Real-World Motion}}\\[0.5em]
        {\large Supplementary Material}
    \end{center}
    \vspace{1em}
}]
\renewcommand{\thesection}{\Alph{section}}
\setcounter{section}{0}

\section{Additional ablations and analyses}

\begin{table}[h]
  \centering
  \captionof{table}{\textbf{Ablation on training video length.}}
  \vspace{-10pt}
  \resizebox{0.75\linewidth}{!}{%
    \begin{tabular}{lc|ccc}
        \toprule
        & \multirow{2}{*}{Video Length} & \multicolumn{3}{c}{DAVIS} \\
        & & AJ & $<\delta^{x}_{avg}$ & OA \\
        \midrule\midrule
        \textbf{(I)}& $24$ & \underline{64.7} & 76.8 & \textbf{89.4} \\
        \textbf{(II)}& $48$  & \textbf{64.8} & \textbf{77.3} & \underline{89.1} \\
        \textbf{(III)}& $64$ & 64.5 & \underline{77.1} & 88.9 \\
        \bottomrule
    \end{tabular}%
  }
  \label{tab:grid_capof_video_length}
  \vspace{-5pt}
\end{table}

\paragrapht{Ablation on training video length.}
In Table~\ref{tab:grid_capof_video_length}, we investigate the influence of training video length on model performance. For this analysis, we fine-tuned the LocoTrack~\cite{cho2024local} base model using our annotated videos, experimenting with distinct clip lengths. The results demonstrate that utilizing video clips of 48 frames yields the best performance in terms of both AJ and $<\delta^{x}_{avg}$.

\begin{table}[h]
  \centering
  \captionof{table}{\textbf{Ablation on frame dilation.}}
  \vspace{-10pt}
    \begin{tabular}{lc|ccc}
        \toprule
        & Frame & \multicolumn{3}{c}{DAVIS} \\
        & Dilation & AJ & $<\delta^{x}_{avg}$ & OA \\
        \midrule\midrule
        \textbf{(I)}& 1 & \underline{64.6} & 76.7 & \textbf{89.6}\\
        \textbf{(II)}& 2  & \textbf{64.8} & \textbf{77.3} & \underline{89.1} \\
        \textbf{(III)}& 3 & 64.2 & \underline{76.9} & 89.0\\
        \bottomrule
    \end{tabular}%
  \label{tab:grid_capof_dilation}
  \vspace{-5pt}
\end{table}

\paragrapht{Ablation on frame dilation.}
During training, we adjust the frame rate to control the motion speed by sampling video frames with different dilation factors. In Table~\ref{tab:grid_capof_dilation}, we test dilations of 1, 2, and 3, corresponding to 1$\times$, 2$\times$, and 3$\times$ faster motion, respectively. When the dilation is too large, the gap between adjacent frames becomes too wide. We found that a dilation factor of 2 yields the best performance in both AJ and $<\delta^{x}_{avg}$.

\begin{table}[h]
  \centering
  \captionof{table}{\textbf{Ablation on using different amount of extra data.}}
  \vspace{-10pt}
  \resizebox{\linewidth}{!}{
    \begin{tabular}{l|ccc}
    \toprule
    \multirow{2}{*}{Method} & \multicolumn{3}{c}{\textbf{DAVIS}} \\
    & AJ &$<\delta^{x}_{avg}$ & OA \\
    
    \midrule\midrule
    
    LocoTrack ($384\times 512$) & 64.8 & 77.4 & 86.2\\
    LocoTrack + \textbf{Ours} (250 videos) & 65.6 & 78.5 & \textbf{87.3}\\
    LocoTrack + \textbf{Ours} (500 videos) & \underline{65.7} & \underline{78.7} & \underline{87.2}\\
    LocoTrack + \textbf{Ours} (full) & \textbf{65.9} & \textbf{78.9} & \textbf{87.3}\\
    
    \bottomrule
    \end{tabular}%
  }
  \label{tab:abl_amout_of_extra_data}
  \vspace{-5pt}
\end{table}

\paragrapht{Ablation on training data scale.}
In Table~\ref{tab:abl_amout_of_extra_data}, we examine how the amount of additional training data influences model performance. Starting from our full dataset containing 1,400 videos, we evaluate reduced subsets of 500 and 250 videos. The results show that even when using only a fraction of the full dataset, our additional videos consistently improve performance over the LocoTrack baseline across all metrics. This demonstrates that our dataset provides strong supervision signals, and meaningful gains can be achieved even with limited data.

\section{Additional visualization}

\begin{figure*}[p]
    \centering
    \begin{minipage}{0.8\textwidth}
        \centering
        \includegraphics[width=\linewidth]{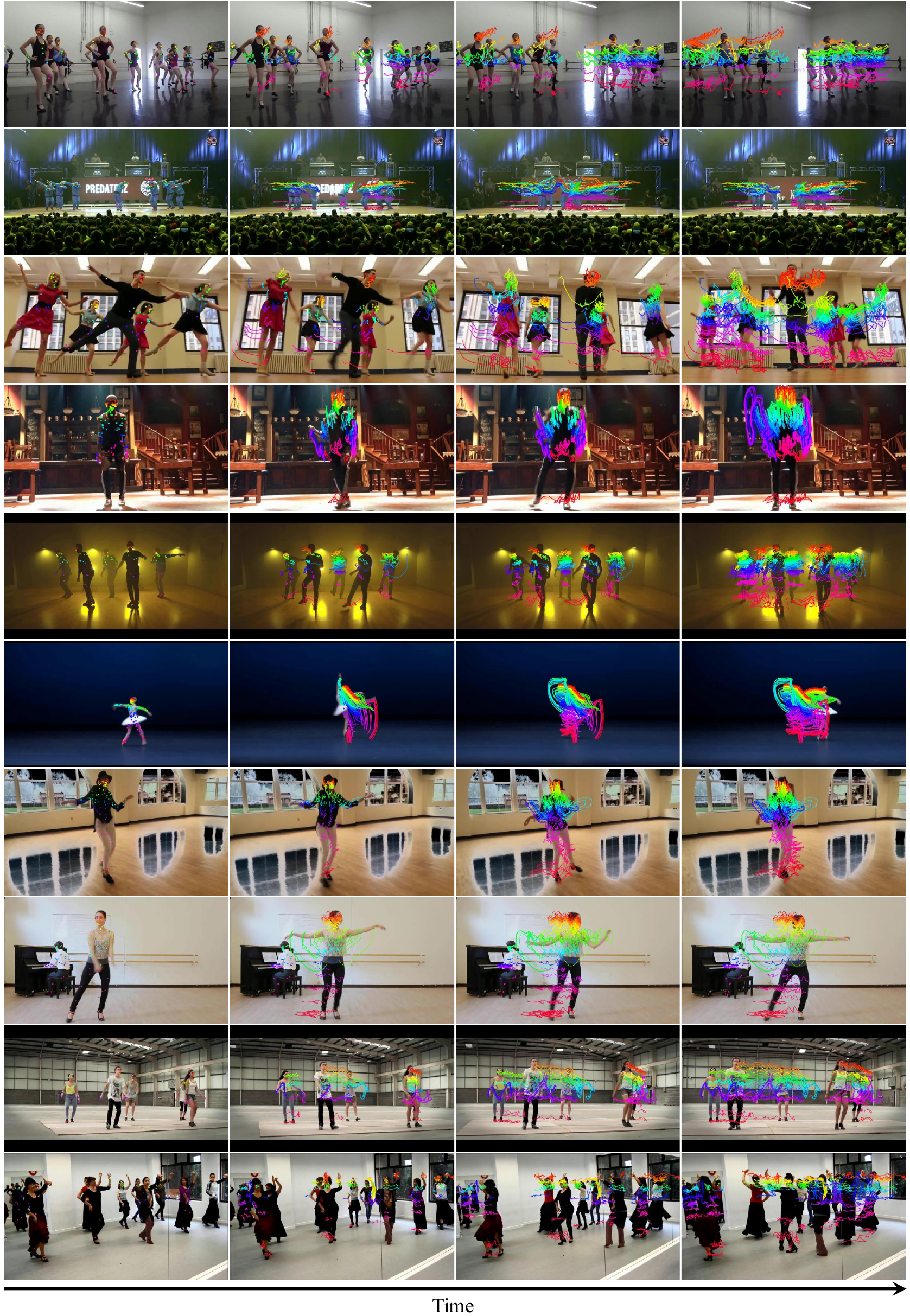}
        \vspace{-15pt}
        \caption{\textbf{Additional visualization of videos annotated with our pipeline.} In this example, we visualize trajectories extracted using our pipeline.}
        \label{fig:supple-video-visualization1}
        \vspace{-10pt}
    \end{minipage}
\end{figure*}

\begin{figure*}[p]
    \centering
    \begin{minipage}{0.8\textwidth}
        \centering
        \includegraphics[width=\linewidth]{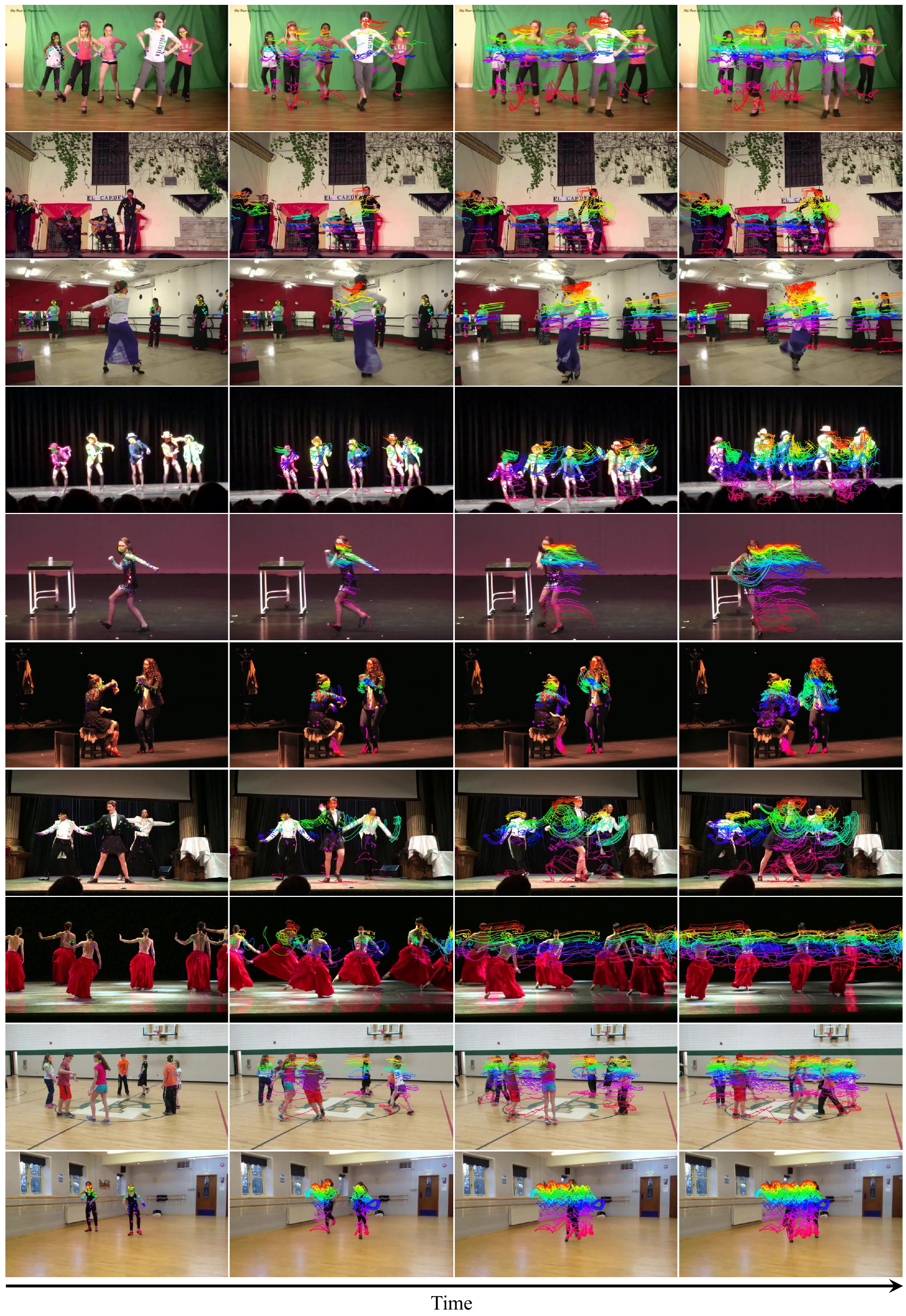}
        \vspace{-15pt}
        \caption{\textbf{Additional visualization of videos annotated with our pipeline.} In this example, we visualize trajectories extracted using our pipeline.}
        \label{fig:supple-video-visualization2}
        \vspace{-10pt}
    \end{minipage}
\end{figure*}

We visualize videos annotated using our pipeline in Figure~\ref{fig:supple-video-visualization1} and Figure~\ref{fig:supple-video-visualization2}. The visualizations highlight the complex motion present in the dataset.

\section{Filtering outlier trajectories with optical flow}
\begin{figure*}[t]
    \centering
    \includegraphics[width=\textwidth]{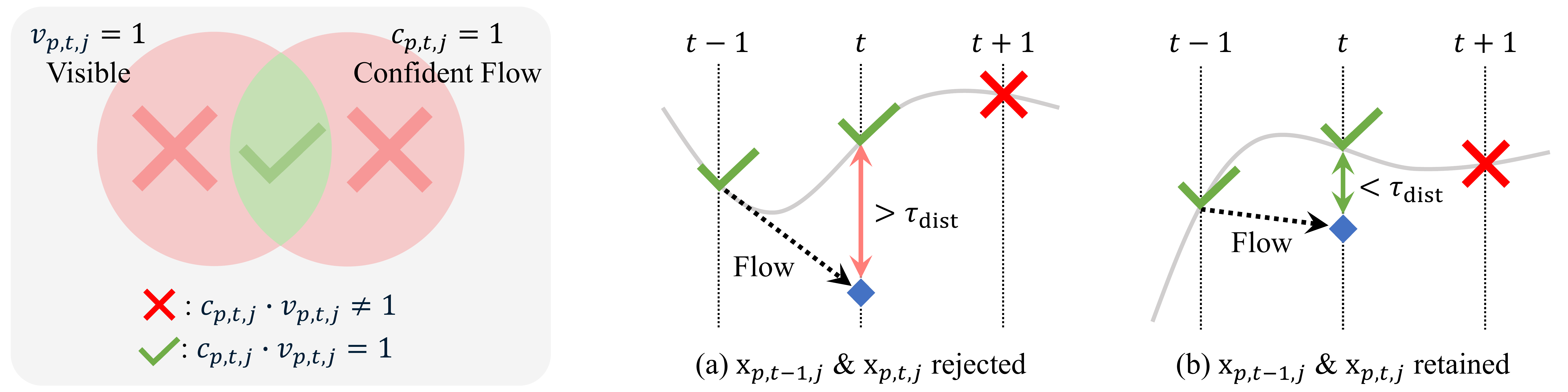}
    \vspace{-15pt}
    \caption{\textbf{Filtering erroneous tracks with optical flow.} We filter trajectories predicted from the human mesh using optical flow. First, we retain points that are both considered visible with ray-casting and have confident optical flow at the predicted position ($c_{p,t,j} \cdot v_{p,t,j} = 1$), denoted as \raisebox{-0.2em}{\includegraphics[height=1em]{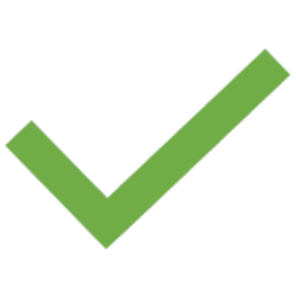}}. If the difference between the trajectory predicted from the human mesh and the optical flow exceeds a threshold $\tau_{\text{dist}}$, the point is considered erroneous, as depicted in (a), otherwise retained as in (b). We omit the normalization process for clarity.}
    \label{fig:filtering}
    \vspace{-10pt}
\end{figure*}

While HMR methods~\cite{black2023bedlam,dwivedi2024tokenhmr,goel2023humans} have become increasingly robust to challenges such as motion blur, rapid movements, and varying lighting conditions due to their strong human priors, they may still produce occasional errors or inconsistencies in mesh predictions under extreme conditions. Furthermore, our ray-casting-based visibility prediction is limited to occlusions caused by other modeled humans and does not account for occlusions from general scene objects. In addition, the parametric nature of the SMPL model makes it less effective at capturing the motion of highly deformable or loosely attached clothing and accessories.

To mitigate the impact of these potential inaccuracies on our pseudo-labels and thus improve the quality of training data, we introduce an optical flow-based filtering stage, illustrated in Figure~\ref{fig:filtering}. Optical flow~\cite{teed2020raft,huang2022flowformer,wang2024sea} is particularly well-suited for this task due to its established accuracy in estimating dense motion between adjacent frames. Over very short temporal windows, these flow-based predictions can provide strong local motion cues to validate or identify discrepancies in the HMR-derived trajectories, especially when dealing with complex non-rigid deformations or potential tracking drift~\cite{cho2024flowtrack,Neoral_2024_WACV}. 

The primary objective of this stage is not to replace the HMR-derived tracks but to identify and remove potentially erroneous segments or entire trajectories. Our goal is to achieve a high true positive rate for valid track segments while maintaining a sufficiently high true negative rate for incorrect ones. We operate under the assumption that training a point tracker with a smaller set of highly accurate pseudo-labels is more beneficial than using a larger set contaminated with significant errors.

\paragrapht{Identify confident optical flow.} For each pair of consecutive frames $(I_t, I_{t+1})$, we compute both the forward optical flow map from $I_t$ to $I_{t+1}$, denoted $\mathbf{F}_{t \to t+1}$, and the backward optical flow map $\mathbf{F}_{t+1 \to t}$. We use $\vect{f}(\mathbf{F}, \vect{x})$ to denote the process of sampling the flow vector from map $\mathbf{F}$ at sub-pixel location $\vect{x}$ using bilinear interpolation. For notational simplicity, we define:
\begin{align}
    \boldsymbol{\xi}_{p,t,j} &= \vect{f}(\mathbf{F}_{t \to t+1}, \vect{x}_{p,t,j}),\quad \\
    \boldsymbol{\zeta}_{p,t,j} &= \vect{f}(\mathbf{F}_{t+1 \to t}, \vect{x}_{p,t,j} + \boldsymbol{\xi}_{p,t,j}).
\end{align}
To assess the reliability of the optical flow estimation itself at a given point $\vect{x}_{p,t,j}$, we apply a forward-backward consistency check~\cite{meister2018unflow}. Let $c_{p,t,j}$ denote the binary indicator of optical flow reliability at point $\vect{x}_{p,t,j}$ in frame $I_t$. A point $\vect{x}_{p,t,j}$ is considered to have reliable flow ($c_{p,t,j}=1$) if the $L_2$ distance between the original point location and the location obtained after warping to frame $t+1$ and back to frame $t$ is below a predefined threshold $\delta_{\text{cons}}$:
\begin{align}
c_{p,t,j} = \mathbbm{1}\left[ \| \boldsymbol{\xi}_{p,t,j} + \boldsymbol{\zeta}_{p,t,j}\|_2 < \delta_{\text{cons}} \right].
\end{align}

\paragrapht{Find erroneous trajectories.} For each point $\vect{x}_{p,t,j}$ within a pseudo-labeled trajectory $\mathcal{X}_{p,j}$ (where $t$ and $t+1$ are both in $\mathcal{T}_{p,j}$), we examine the predicted motion to the next frame, $\vect{x}_{p,t+1,j}$. The displacement vector derived from the HMR pseudo-label is $\Delta \vect{x}_{p,t,j}^{\text{HMR}} = \vect{x}_{p,t+1,j} - \vect{x}_{p,t,j}$. We compare this to the optical flow displacement $\boldsymbol{\xi}_{p,t,j}$ (previously defined as the forward flow at location $\vect{x}_{p,t,j}$).

To compare these two displacement vectors, $\Delta \vect{x}_{p,t,j}^{\text{HMR}}$ and $\boldsymbol{\xi}_{p,t,j}$, robustly, especially in cases of large motion, we normalize them. Let $L_{\text{short}} = \min(\|\Delta \vect{x}_{p,t,j}^{\text{HMR}}\|_{2}, \|\boldsymbol{\xi}_{p,t,j}\|_{2}) + \epsilon_{\text{norm}}$, where $\epsilon_{\text{norm}}$ is a small positive constant added to prevent division by zero. The normalized vectors are:
\begin{equation}
\Delta \widehat{\vect{x}}_{p,t,j}^{\text{HMR}} = \frac{\Delta \vect{x}_{p,t,j}^{\text{HMR}}}{L_{\text{short}}}, \quad
\widehat{\boldsymbol{\xi}}_{p,t,j} = \frac{\boldsymbol{\xi}_{p,t,j}}{L_{\text{short}}}.
\end{equation}
A point transition from frame $t$ to $t+1$ is flagged as potentially erroneous if the $L_2$ distance between these normalized displacement vectors exceeds a threshold $\tau_{\text{dist}}$. We define an indicator variable $e_{p,t,j}$ for this:
\begin{equation}
e_{p,t,j} = \mathbbm{1} \left[ \| \Delta \widehat{\vect{x}}_{p,t,j}^{\text{HMR}} - \widehat{\boldsymbol{\xi}}_{p,t,j} \|_{2} > \tau_{\text{dist}} \right].
\end{equation}

\paragrapht{Trajectory rejection.} We observed that query points located on regions not well captured by the SMPL model, such as loose clothing or hair, often result in a large number of transitions being flagged as erroneous. Nonetheless, even in these cases, certain frames may produce transitions that closely follow the predicted motion and are not flagged as erroneous. To robustly filter such trajectories, we evaluate the proportion of transitions that are flagged as erroneous. For a given trajectory $\mathcal{X}_{p,j}$ (defined by the ordered sequence of frame indices $\mathcal{T}_{p,j}$), we calculate the fraction of its transitions that are flagged as erroneous. This ratio, $R_{p,j}$, is computed by summing over all transitions from a frame $t_k$ to its successor $t_{k+1}$ within the sequence $\mathcal{T}_{p,j}$. Only transitions where the point at the starting frame $t_k$ is estimated as visible ($v_{p,t_k,j}=1$) by ray casting and the optical flow is deemed reliable ($c_{p,t_k,j}=1$) are included in this calculation:
\begin{equation}
R_{p,j} = \frac{\sum_{k=1}^{|\mathcal{T}_{p,j}|-1} e_{p,t_k,j} \cdot v_{p,t_k,j} \cdot c_{p,t_k,j}}{\sum_{k=1}^{|\mathcal{T}_{p,j}|-1} v_{p,t_k,j} \cdot c_{p,t_k,j} + \epsilon_{\text{ratio}}},
\end{equation}
where $(t_k, t_{k+1})$ denotes the $k$-th pair of consecutive frame indices in the ordered set $\mathcal{T}_{p,j}$. The term $e_{p,t_k,j}$ indicates that the transition starting at frame $t_k$ (and ending at $t_{k+1}$) is erroneous. $\epsilon_{\text{ratio}}$ is a small constant to ensure numerical stability in case the denominator is zero. A trajectory $\mathcal{X}_{p,j}$ is ultimately rejected if this erroneous transition ratio $R_{p,j}$ exceeds a predefined threshold $\tau_{\text{ratio}}$.

For trajectories that pass this trajectory-level filtering, any individual point-pair transition $(\vect{x}_{p,t,j}, \vect{x}_{p,t+1,j})$ still flagged as erroneous ($e_{p,t,j}=1$) is considered unreliable. For the purpose of training a point tracking model, such individual erroneous transitions are excluded.

\section{Trajectory complexity and diversity analysis}
To quantitatively evaluate the characteristics of generated pseudo-labeled trajectories $\mathcal{X}_{p,j}$, we employ metrics for trajectory complexity and dataset diversity. These metrics help in understanding the nature of the motion patterns captured. All trajectories are sequences of 2D points $\vect{x}_{p,t,j}$ over time $t$. Calculations are performed on contiguous visible segments of these trajectories, and points are assumed to be normalized by frame dimensions.

\paragrapht{Trajectory complexity.}
Trajectory complexity is quantified using the mean angular acceleration magnitude. This metric captures the rate of change in the direction of motion, highlighting non-linear movements and directional variations. For a given visible segment of a trajectory $\mathcal{X}_{p,j}$, consisting of an ordered sequence of points $(\vect{x}_0, \vect{x}_1, \ldots, \vect{x}_K)$ (where $\vect{x}_k$ corresponds to a point $\vect{x}_{p,t_k,j}$ from the trajectory, requiring at least 4 points with $K \ge 3$ for at least one angular acceleration value), and assuming a constant time step $\Delta t$ between frames:

First, a sequence of velocity vectors $\vect{u}_k$ between consecutive points is computed:
\begin{align}
\vect{u}_k = \vect{x}_{k+1} - \vect{x}_k \quad \text{for } k = 0, \ldots, K-1.
\end{align}
The signed turning angle $\theta_k$ at point $\vect{x}_{k+1}$ (i.e., between vectors $\vect{u}_k$ and $\vect{u}_{k+1}$) is calculated as:
\begin{equation}
\theta_k = \text{atan2}(\vect{u}_k^{(1)}\vect{u}_{k+1}^{(2)} - \vect{u}_k^{(2)}\vect{u}_{k+1}^{(1)}, \vect{u}_k \cdot \vect{u}_{k+1}),
\end{equation}
for $k = 0, \ldots, K-2$.
This sequence of angles is then unwrapped to obtain a continuous sequence of angles $\Phi = (\phi_0, \phi_1, \ldots, \phi_{K-2})$ by adding multiples of $\pm 2\pi$ to eliminate jumps.
The sequence of angular velocities $\omega_k$ is computed from the unwrapped turning angles:
\begin{align}
\omega_k = \frac{\phi_k}{\Delta t} \quad \text{for } k = 0, \ldots, K-2.
\end{align}
The angular accelerations $\alpha_k$ are then found by taking the difference between consecutive angular velocities:
\begin{align}
\alpha_k = \frac{\omega_{k+1} - \omega_k}{\Delta t} \quad \text{for } k = 0, \ldots, K-3.
\end{align}
The complexity for the segment, $C_{\text{seg}}$, is the mean of the magnitudes of these angular accelerations:
\begin{align}
C_{\text{seg}} = \frac{1}{K-2} \sum_{k=0}^{K-3} |\alpha_k|.
\end{align}
The complexity for an entire trajectory $\mathcal{X}_{p,j}$, denoted $C_{\mathcal{X}_{p,j}}$, is the average of $C_{\text{seg}}$ over all its valid visible segments. The overall dataset complexity is the mean of $C_{\mathcal{X}_{p,j}}$ across all trajectories.

\paragrapht{Trajectory diversity.}
Trajectory diversity is assessed by computing the mean standard deviation of centered trajectories from a mean centered trajectory. This measures the spatial variability of trajectories in terms of their shape and relative motion, independent of their absolute starting positions.

Consider a set of $N$ trajectories $\{\mathcal{X}_1, \mathcal{X}_2, \ldots, \mathcal{X}_N\}$ within a single video, where $\mathcal{X}_i$ refers to a specific $\mathcal{X}_{p,j}$, and $\mathcal{X}_i(t)$ denotes the 2D coordinate $\vect{x}_{p,t,j}$ for the $i$-th trajectory at frame $t$. First, each trajectory $\mathcal{X}_i(t)$ is centered by subtracting the coordinates of its first visible point $\mathcal{X}_i(t_{i, \text{first\_vis}})$ from all its subsequent visible points. Let this centered trajectory be $\mathcal{X}'_i(t)$:
\begin{align}
\mathcal{X}'_i(t) = \mathcal{X}_i(t) - \mathcal{X}_i(t_{i, \text{first\_vis}})\quad \text{for visible } t \ge t_{i, \text{first\_vis}}.
\end{align}
Points where the trajectory is occluded or prior to $t_{i, \text{first\_vis}}$ are considered invalid.
A mean centered trajectory $M'(t)$ is computed by averaging the coordinates of all valid centered trajectories at each frame $t$:
\begin{align}
M'(t) = \frac{1}{|\mathcal{V}_t|} \sum_{i \in \mathcal{V}_t} \mathcal{X}'_i(t),
\end{align}
where $\mathcal{V}_t$ is the set of indices of trajectories having valid centered data at frame $t$.
For each centered trajectory $\mathcal{X}'_i$, the mean squared Euclidean distance ($MSD_i$) from $M'(t)$ is calculated over all frames $t \in \mathcal{T}_i$ where both $\mathcal{X}'_i(t)$ and $M'(t)$ are valid:
\begin{align}
MSD_i = \frac{1}{|\mathcal{T}_i|} \sum_{t \in \mathcal{T}_i} \|\mathcal{X}'_i(t) - M'(t)\|_2^2.
\end{align}
The standard deviation for trajectory $i$ is then $SD_i = \sqrt{MSD_i}$.
The diversity score $D_{\text{video}}$ for the video (or dataset) is the mean of these individual trajectory standard deviations:
\begin{align}
D_{\text{video}} = \text{mean}(SD_1, SD_2, \ldots, SD_N).
\end{align}

\begin{figure}[t]
    \centering
    \includegraphics[width=0.5\linewidth]{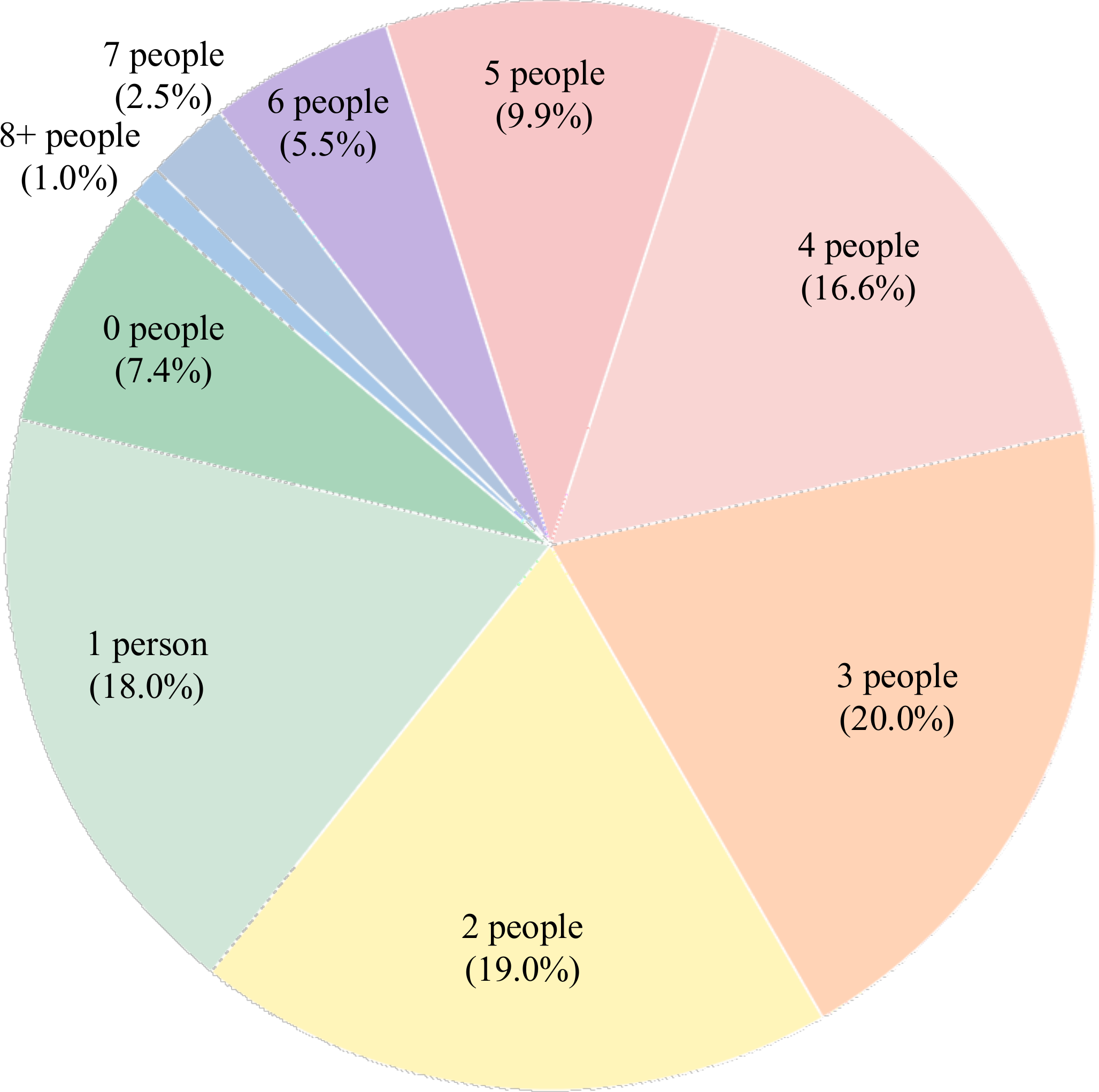}
    \caption{\textbf{The number of people per frame.} Anthro-LD, the dataset generated with our pipeline, contains videos where most scenes include more than one person, contributing to the complexity of the trajectories.}
    \label{fig:supp_stats}
    \vspace{-10pt}
\end{figure}

\section{Additional data statistics}
\paragrapht{Statistics on the number of people in training videos.}
In Figure~\ref{fig:supp_stats}, we investigate the number of people per frame in the generated dataset. Most frames contain more than one person (82\%), which can lead to complex trajectories due to occlusions between individuals.

\section{Limitations and social impact}
While our pseudo-labels provide substantial performance gains when used to train point tracking models, the filtering pipeline is designed conservatively to remove erroneous trajectories, which means some potentially valid trajectories may be rejected. Furthermore, visible points in the video could be mistakenly identified as occluded. This inaccuracy regarding occlusion status led us to decide against supervising occlusion directly, and instead focus only on position. Although a dataset with inaccurate occlusion information can still be helpful for training, its utility as a benchmark is limited. 

\paragrapht{Social impact.} This work significantly enhances point tracking accessibility and efficiency by automating data generation, thereby reducing reliance on costly manual annotation and extensive computational resources. This advancement can benefit robotics, 3D/4D reconstruction, and video editing. Open-sourcing the dataset and pipeline will promote collaborative research on point tracking. While not explicitly discussed, potential misuse for surveillance should be considered.

\section*{Acknowledgment}
This research was supported by Institute of Information \& communications Technology Planning \& Evaluation (IITP) grant funded by the Korea government (MSIT) (RS-2019-II190075, RS-2024-00509279, RS-2025-II212068, RS-2023-00227592, RS-2025- 02214479, RS-2024-00457882, RS-2025-25441838, RS-2025-25441838, RS-2025-02214479, RS-2025-02217259) and the Culture, Sports, and Tourism R\&D Program through the Korea Creative Content Agency grant funded by the Ministry of Culture, Sports and Tourism (RS-2024-00345025, RS-2024-00333068, RS-2023-00222280, RS-2023-00266509), and National Research Foundation of Korea (RS-2024-00346597).

\clearpage

\clearpage

{
    \small
    \bibliographystyle{ieeenat_fullname}
    \bibliography{main}

@String(CVPR= {IEEE Conf. Comput. Vis. Pattern Recog.})

@String(ECCV= {Eur. Conf. Comput. Vis.})

@String(AAAI = {AAAI})

@String(CVPR  = {CVPR})

@String(ECCV  = {ECCV})

@inproceedings{greff2022kubric,
  title={Kubric: A scalable dataset generator},
  author={Greff, Klaus and Belletti, Francois and Beyer, Lucas and Doersch, Carl and Du, Yilun and Duckworth, Daniel and Fleet, David J and Gnanapragasam, Dan and Golemo, Florian and Herrmann, Charles and others},
  booktitle={Proceedings of the IEEE/CVF conference on computer vision and pattern recognition},
  pages={3749--3761},
  year={2022}
}

@article{kay2017kinetics,
  title={The kinetics human action video dataset},
  author={Kay, Will and Carreira, Joao and Simonyan, Karen and Zhang, Brian and Hillier, Chloe and Vijayanarasimhan, Sudheendra and Viola, Fabio and Green, Tim and Back, Trevor and Natsev, Paul and others},
  journal={arXiv preprint arXiv:1705.06950},
  year={2017}
}

@article{pont20172017,
  title={The 2017 davis challenge on video object segmentation},
  author={Pont-Tuset, Jordi and Perazzi, Federico and Caelles, Sergi and Arbel{\'a}ez, Pablo and Sorkine-Hornung, Alex and Van Gool, Luc},
  journal={arXiv preprint arXiv:1704.00675},
  year={2017}
}

@inproceedings{zheng2023pointodyssey,
  title={Pointodyssey: A large-scale synthetic dataset for long-term point tracking},
  author={Zheng, Yang and Harley, Adam W and Shen, Bokui and Wetzstein, Gordon and Guibas, Leonidas J},
  booktitle={Proceedings of the IEEE/CVF International Conference on Computer Vision},
  pages={19855--19865},
  year={2023}
}

@inproceedings{teed2020raft,
  title={Raft: Recurrent all-pairs field transforms for optical flow},
  author={Teed, Zachary and Deng, Jia},
  booktitle={Computer Vision--ECCV 2020: 16th European Conference, Glasgow, UK, August 23--28, 2020, Proceedings, Part II 16},
  pages={402--419},
  year={2020},
  organization={Springer}
}

@inproceedings{harley2022particle,
  title={Particle video revisited: Tracking through occlusions using point trajectories},
  author={Harley, Adam W and Fang, Zhaoyuan and Fragkiadaki, Katerina},
  booktitle={European Conference on Computer Vision},
  pages={59--75},
  year={2022},
  organization={Springer}
}

@article{doersch2022tap,
  title={Tap-vid: A benchmark for tracking any point in a video},
  author={Doersch, Carl and Gupta, Ankush and Markeeva, Larisa and Recasens, Adria and Smaira, Lucas and Aytar, Yusuf and Carreira, Joao and Zisserman, Andrew and Yang, Yi},
  journal={Advances in Neural Information Processing Systems},
  volume={35},
  pages={13610--13626},
  year={2022}
}

@inproceedings{doersch2023tapir,
  title={Tapir: Tracking any point with per-frame initialization and temporal refinement},
  author={Doersch, Carl and Yang, Yi and Vecerik, Mel and Gokay, Dilara and Gupta, Ankush and Aytar, Yusuf and Carreira, Joao and Zisserman, Andrew},
  booktitle={Proceedings of the IEEE/CVF International Conference on Computer Vision},
  pages={10061--10072},
  year={2023}
}

@inproceedings{karaev2024cotracker,
  title={Cotracker: It is better to track together},
  author={Karaev, Nikita and Rocco, Ignacio and Graham, Benjamin and Neverova, Natalia and Vedaldi, Andrea and Rupprecht, Christian},
  booktitle={European Conference on Computer Vision},
  pages={18--35},
  year={2024},
  organization={Springer}
}

@inproceedings{cho2024local,
  title={Local All-Pair Correspondence for Point Tracking},
  author={Cho, Seokju and Huang, Jiahui and Nam, Jisu and An, Honggyu and Kim, Seungryong and Lee, Joon-Young},
  booktitle={European Conference on Computer Vision},
  pages={306--325},
  year={2024},
  organization={Springer}
}

@article{karaev2024cotracker3,
  title={CoTracker3: Simpler and better point tracking by pseudo-labelling real videos},
  author={Karaev, Nikita and Makarov, Iurii and Wang, Jianyuan and Neverova, Natalia and Vedaldi, Andrea and Rupprecht, Christian},
  journal={arXiv preprint arXiv:2410.11831},
  year={2024}
}

@inproceedings{doersch2024bootstap,
  title={Bootstap: Bootstrapped training for tracking-any-point},
  author={Doersch, Carl and Luc, Pauline and Yang, Yi and Gokay, Dilara and Koppula, Skanda and Gupta, Ankush and Heyward, Joseph and Rocco, Ignacio and Goroshin, Ross and Carreira, Jo{\~a}o and others},
  booktitle={Proceedings of the Asian Conference on Computer Vision},
  pages={3257--3274},
  year={2024}
}

@article{kim2025exploring,
  title={Exploring Temporally-Aware Features for Point Tracking},
  author={Kim, In{\`e}s Hyeonsu and Cho, Seokju and Huang, Jiahui and Yi, Jung and Lee, Joon-Young and Kim, Seungryong},
  journal={arXiv preprint arXiv:2501.12218},
  year={2025}
}

@inproceedings{wang2025vggt,
  title={VGGT: Visual Geometry Grounded Transformer},
  author={Wang, Jianyuan and Chen, Minghao and Karaev, Nikita and Vedaldi, Andrea and Rupprecht, Christian and Novotny, David},
  booktitle={Proceedings of the IEEE/CVF Conference on Computer Vision and Pattern Recognition},
  year={2025}
}

@inproceedings{balasingam2024drivetrack,
  title={Drivetrack: A benchmark for long-range point tracking in real-world videos},
  author={Balasingam, Arjun and Chandler, Joseph and Li, Chenning and Zhang, Zhoutong and Balakrishnan, Hari},
  booktitle={Proceedings of the IEEE/CVF Conference on Computer Vision and Pattern Recognition},
  pages={22488--22497},
  year={2024}
}

@inproceedings{lee2021beyond,
  title={Beyond pick-and-place: Tackling robotic stacking of diverse shapes},
  author={Lee, Alex X and Devin, Coline Manon and Zhou, Yuxiang and Lampe, Thomas and Bousmalis, Konstantinos and Springenberg, Jost Tobias and Byravan, Arunkumar and Abdolmaleki, Abbas and Gileadi, Nimrod and Khosid, David and others},
  booktitle={5th Annual Conference on Robot Learning},
  year={2021}
}

@inproceedings{karaev2023dynamicstereo,
  title={Dynamicstereo: Consistent dynamic depth from stereo videos},
  author={Karaev, Nikita and Rocco, Ignacio and Graham, Benjamin and Neverova, Natalia and Vedaldi, Andrea and Rupprecht, Christian},
  booktitle={Proceedings of the IEEE/CVF Conference on Computer Vision and Pattern Recognition},
  pages={13229--13239},
  year={2023}
}

@inproceedings{loper2023smpl,
  title={SMPL: A skinned multi-person linear model},
  author={Loper, Matthew and Mahmood, Naureen and Romero, Javier and Pons-Moll, Gerard and Black, Michael J},
  booktitle={Seminal Graphics Papers: Pushing the Boundaries, Volume 2},
  pages={851--866},
  year={2023}
}

@article{geng2024motion,
  title={Motion prompting: Controlling video generation with motion trajectories},
  author={Geng, Daniel and Herrmann, Charles and Hur, Junhwa and Cole, Forrester and Zhang, Serena and Pfaff, Tobias and Lopez-Guevara, Tatiana and Doersch, Carl and Aytar, Yusuf and Rubinstein, Michael and others},
  journal={arXiv preprint arXiv:2412.02700},
  year={2024}
}

@inproceedings{bogo2016keep,
  title={Keep it SMPL: Automatic estimation of 3D human pose and shape from a single image},
  author={Bogo, Federica and Kanazawa, Angjoo and Lassner, Christoph and Gehler, Peter and Romero, Javier and Black, Michael J},
  booktitle={Computer Vision--ECCV 2016: 14th European Conference, Amsterdam, The Netherlands, October 11-14, 2016, Proceedings, Part V 14},
  pages={561--578},
  year={2016},
  organization={Springer}
}

@inproceedings{lassner2017unite,
  title={Unite the people: Closing the loop between 3d and 2d human representations},
  author={Lassner, Christoph and Romero, Javier and Kiefel, Martin and Bogo, Federica and Black, Michael J and Gehler, Peter V},
  booktitle={Proceedings of the IEEE conference on computer vision and pattern recognition},
  pages={6050--6059},
  year={2017}
}

@inproceedings{cao2017realtime,
  title={Realtime multi-person 2d pose estimation using part affinity fields},
  author={Cao, Zhe and Simon, Tomas and Wei, Shih-En and Sheikh, Yaser},
  booktitle={Proceedings of the IEEE conference on computer vision and pattern recognition},
  pages={7291--7299},
  year={2017}
}

@article{xu2022vitpose,
  title={Vitpose: Simple vision transformer baselines for human pose estimation},
  author={Xu, Yufei and Zhang, Jing and Zhang, Qiming and Tao, Dacheng},
  journal={Advances in neural information processing systems},
  volume={35},
  pages={38571--38584},
  year={2022}
}

@inproceedings{goel2023humans,
  title={Humans in 4D: Reconstructing and tracking humans with transformers},
  author={Goel, Shubham and Pavlakos, Georgios and Rajasegaran, Jathushan and Kanazawa, Angjoo and Malik, Jitendra},
  booktitle={Proceedings of the IEEE/CVF International Conference on Computer Vision},
  pages={14783--14794},
  year={2023}
}

@inproceedings{dwivedi2024tokenhmr,
  title={Tokenhmr: Advancing human mesh recovery with a tokenized pose representation},
  author={Dwivedi, Sai Kumar and Sun, Yu and Patel, Priyanka and Feng, Yao and Black, Michael J},
  booktitle={Proceedings of the IEEE/CVF Conference on Computer Vision and Pattern Recognition},
  pages={1323--1333},
  year={2024}
}

@inproceedings{li2024taptr,
  title={Taptr: Tracking any point with transformers as detection},
  author={Li, Hongyang and Zhang, Hao and Liu, Shilong and Zeng, Zhaoyang and Ren, Tianhe and Li, Feng and Zhang, Lei},
  booktitle={European Conference on Computer Vision},
  pages={57--75},
  year={2024},
  organization={Springer}
}

@inproceedings{cho2024flowtrack,
  title={Flowtrack: Revisiting optical flow for long-range dense tracking},
  author={Cho, Seokju and Huang, Jiahui and Kim, Seungryong and Lee, Joon-Young},
  booktitle={Proceedings of the IEEE/CVF Conference on Computer Vision and Pattern Recognition},
  pages={19268--19277},
  year={2024}
}

@article{loshchilov2017decoupled,
  title={Decoupled weight decay regularization},
  author={Loshchilov, Ilya and Hutter, Frank},
  journal={arXiv preprint arXiv:1711.05101},
  year={2017}
}

@article{oquab2023dinov2,
  title={Dinov2: Learning robust visual features without supervision},
  author={Oquab, Maxime and Darcet, Timoth{\'e}e and Moutakanni, Th{\'e}o and Vo, Huy and Szafraniec, Marc and Khalidov, Vasil and Fernandez, Pierre and Haziza, Daniel and Massa, Francisco and El-Nouby, Alaaeldin and others},
  journal={arXiv preprint arXiv:2304.07193},
  year={2023}
}

@inproceedings{sajjadi2022scene,
  title={Scene representation transformer: Geometry-free novel view synthesis through set-latent scene representations},
  author={Sajjadi, Mehdi SM and Meyer, Henning and Pot, Etienne and Bergmann, Urs and Greff, Klaus and Radwan, Noha and Vora, Suhani and Lu{\v{c}}i{\'c}, Mario and Duckworth, Daniel and Dosovitskiy, Alexey and others},
  booktitle={Proceedings of the IEEE/CVF Conference on Computer Vision and Pattern Recognition},
  pages={6229--6238},
  year={2022}
}

@article{singh2022simple,
  title={Simple unsupervised object-centric learning for complex and naturalistic videos},
  author={Singh, Gautam and Wu, Yi-Fu and Ahn, Sungjin},
  journal={Advances in Neural Information Processing Systems},
  volume={35},
  pages={18181--18196},
  year={2022}
}

@article{ren2025gen3c,
  title={Gen3c: 3d-informed world-consistent video generation with precise camera control},
  author={Ren, Xuanchi and Shen, Tianchang and Huang, Jiahui and Ling, Huan and Lu, Yifan and Nimier-David, Merlin and M{\"u}ller, Thomas and Keller, Alexander and Fidler, Sanja and Gao, Jun},
  journal={arXiv preprint arXiv:2503.03751},
  year={2025}
}

@inproceedings{wang2023tracking,
  title={Tracking everything everywhere all at once},
  author={Wang, Qianqian and Chang, Yen-Yu and Cai, Ruojin and Li, Zhengqi and Hariharan, Bharath and Holynski, Aleksander and Snavely, Noah},
  booktitle={Proceedings of the IEEE/CVF International Conference on Computer Vision},
  pages={19795--19806},
  year={2023}
}

@inproceedings{tumanyan2024dino,
  title={Dino-tracker: Taming dino for self-supervised point tracking in a single video},
  author={Tumanyan, Narek and Singer, Assaf and Bagon, Shai and Dekel, Tali},
  booktitle={European Conference on Computer Vision},
  pages={367--385},
  year={2024},
  organization={Springer}
}

@article{aydemir2025track,
  title={Track-On: Transformer-based Online Point Tracking with Memory},
  author={Aydemir, G{\"o}rkay and Cai, Xiongyi and Xie, Weidi and G{\"u}ney, Fatma},
  journal={arXiv preprint arXiv:2501.18487},
  year={2025}
}

@article{li2024taptrv2,
  title={Taptrv2: Attention-based position update improves tracking any point},
  author={Li, Hongyang and Zhang, Hao and Liu, Shilong and Zeng, Zhaoyang and Li, Feng and Li, Bohan and Ren, Tianhe and Zhang, Lei},
  journal={Advances in Neural Information Processing Systems},
  volume={37},
  pages={101074--101095},
  year={2024}
}

@article{qu2024taptrv3,
  title={TAPTRv3: Spatial and Temporal Context Foster Robust Tracking of Any Point in Long Video},
  author={Qu, Jinyuan and Li, Hongyang and Liu, Shilong and Ren, Tianhe and Zeng, Zhaoyang and Zhang, Lei},
  journal={arXiv preprint arXiv:2411.18671},
  year={2024}
}

@InProceedings{Neoral_2024_WACV,
    author    = {Neoral, Michal and \v{S}er\'ych, Jon\'a\v{s} and Matas, Ji\v{r}{\'\i}},
    title     = {MFT: Long-Term Tracking of Every Pixel},
    booktitle = {Proceedings of the IEEE/CVF Winter Conference on Applications of Computer Vision (WACV)},
    month     = {January},
    year      = {2024},
    pages     = {6837-6847}
}

@incollection{moller2005fast,
  title={Fast, minimum storage ray/triangle intersection},
  author={M{\"o}ller, Tomas and Trumbore, Ben},
  booktitle={ACM SIGGRAPH 2005 Courses},
  pages={7--es},
  year={2005}
}

@inproceedings{black2023bedlam,
  title={Bedlam: A synthetic dataset of bodies exhibiting detailed lifelike animated motion},
  author={Black, Michael J and Patel, Priyanka and Tesch, Joachim and Yang, Jinlong},
  booktitle={Proceedings of the IEEE/CVF Conference on Computer Vision and Pattern Recognition},
  pages={8726--8737},
  year={2023}
}

@inproceedings{huang2022flowformer,
  title={Flowformer: A transformer architecture for optical flow},
  author={Huang, Zhaoyang and Shi, Xiaoyu and Zhang, Chao and Wang, Qiang and Cheung, Ka Chun and Qin, Hongwei and Dai, Jifeng and Li, Hongsheng},
  booktitle={European conference on computer vision},
  pages={668--685},
  year={2022},
  organization={Springer}
}

@inproceedings{wang2024sea,
  title={Sea-raft: Simple, efficient, accurate raft for optical flow},
  author={Wang, Yihan and Lipson, Lahav and Deng, Jia},
  booktitle={European Conference on Computer Vision},
  pages={36--54},
  year={2024},
  organization={Springer}
}

@article{SMPL:2015,
      author = {Loper, Matthew and Mahmood, Naureen and Romero, Javier and Pons-Moll, Gerard and Black, Michael J.},
      title = {{SMPL}: A Skinned Multi-Person Linear Model},
      journal = {ACM Trans. Graphics (Proc. SIGGRAPH Asia)},
      month = oct,
      number = {6},
      pages = {248:1--248:16},
      publisher = {ACM},
      volume = {34},
      year = {2015}
    }

@inproceedings{meister2018unflow,
  title={Unflow: Unsupervised learning of optical flow with a bidirectional census loss},
  author={Meister, Simon and Hur, Junhwa and Roth, Stefan},
  booktitle={Proceedings of the AAAI conference on artificial intelligence},
  volume={32},
  number={1},
  year={2018}
}

@inproceedings{vecerik2024robotap,
  title={Robotap: Tracking arbitrary points for few-shot visual imitation},
  author={Vecerik, Mel and Doersch, Carl and Yang, Yi and Davchev, Todor and Aytar, Yusuf and Zhou, Guangyao and Hadsell, Raia and Agapito, Lourdes and Scholz, Jon},
  booktitle={2024 IEEE International Conference on Robotics and Automation (ICRA)},
  pages={5397--5403},
  year={2024},
  organization={IEEE}
}

@article{castro2018let,
  title={Let's dance: Learning from online dance videos},
  author={Castro, Daniel and Hickson, Steven and Sangkloy, Patsorn and Mittal, Bhavishya and Dai, Sean and Hays, James and Essa, Irfan},
  journal={arXiv preprint arXiv:1801.07388},
  year={2018}
}

@inproceedings{mayer2016large,
  title={A large dataset to train convolutional networks for disparity, optical flow, and scene flow estimation},
  author={Mayer, Nikolaus and Ilg, Eddy and Hausser, Philip and Fischer, Philipp and Cremers, Daniel and Dosovitskiy, Alexey and Brox, Thomas},
  booktitle={Proceedings of the IEEE conference on computer vision and pattern recognition},
  pages={4040--4048},
  year={2016}
}

@InProceedings{Sun_2020_CVPR,
author = {Sun, Pei and Kretzschmar, Henrik and Dotiwalla, Xerxes and Chouard, Aurelien and Patnaik, Vijaysai and Tsui, Paul and Guo, James and Zhou, Yin and Chai, Yuning and Caine, Benjamin and Vasudevan, Vijay and Han, Wei and Ngiam, Jiquan and Zhao, Hang and Timofeev, Aleksei and Ettinger, Scott and Krivokon, Maxim and Gao, Amy and Joshi, Aditya and Zhang, Yu and Shlens, Jonathon and Chen, Zhifeng and Anguelov, Dragomir},
title = {Scalability in Perception for Autonomous Driving: Waymo Open Dataset},
booktitle = {Proceedings of the IEEE/CVF Conference on Computer Vision and Pattern Recognition (CVPR)},
month = {June},
year = {2020}
}

@inproceedings{chen2024leap,
  title={Leap-vo: Long-term effective any point tracking for visual odometry},
  author={Chen, Weirong and Chen, Le and Wang, Rui and Pollefeys, Marc},
  booktitle={Proceedings of the IEEE/CVF Conference on Computer Vision and Pattern Recognition},
  pages={19844--19853},
  year={2024}
}

@article{yang2025magma,
  title={Magma: A foundation model for multimodal ai agents},
  author={Yang, Jianwei and Tan, Reuben and Wu, Qianhui and Zheng, Ruijie and Peng, Baolin and Liang, Yongyuan and Gu, Yu and Cai, Mu and Ye, Seonghyeon and Jang, Joel and others},
  journal={arXiv preprint arXiv:2502.13130},
  year={2025}
}

@article{wang2024shape,
  title={Shape of motion: 4d reconstruction from a single video},
  author={Wang, Qianqian and Ye, Vickie and Gao, Hang and Austin, Jake and Li, Zhengqi and Kanazawa, Angjoo},
  journal={arXiv preprint arXiv:2407.13764},
  year={2024}
}

@InProceedings{Xiao_2024_CVPR,
    author    = {Xiao, Yuxi and Wang, Qianqian and Zhang, Shangzhan and Xue, Nan and Peng, Sida and Shen, Yujun and Zhou, Xiaowei},
    title     = {SpatialTracker: Tracking Any 2D Pixels in 3D Space},
    booktitle = {Proceedings of the IEEE/CVF Conference on Computer Vision and Pattern Recognition (CVPR)},
    month     = {June},
    year      = {2024},
    pages     = {20406-20417}
}

@inproceedings{bharadhwaj2024track2act,
  title={Track2act: Predicting point tracks from internet videos enables generalizable robot manipulation},
  author={Bharadhwaj, Homanga and Mottaghi, Roozbeh and Gupta, Abhinav and Tulsiani, Shubham},
  booktitle={European Conference on Computer Vision},
  pages={306--324},
  year={2024},
  organization={Springer}
}

@article{cho2025seurat,
  title={Seurat: From Moving Points to Depth},
  author={Cho, Seokju and Huang, Jiahui and Kim, Seungryong and Lee, Joon-Young},
  journal={arXiv preprint arXiv:2504.14687},
  year={2025}
}

@article{zhang2025tapip3d,
  title={TAPIP3D: Tracking Any Point in Persistent 3D Geometry},
  author={Zhang, Bowei and Ke, Lei and Harley, Adam W and Fragkiadaki, Katerina},
  journal={arXiv preprint arXiv:2504.14717},
  year={2025}
}

@article{seidenschwarz2024dynomo,
  title={Dynomo: Online point tracking by dynamic online monocular gaussian reconstruction},
  author={Seidenschwarz, Jenny and Zhou, Qunjie and Duisterhof, Bardienus and Ramanan, Deva and Leal-Taix{\'e}, Laura},
  journal={arXiv preprint arXiv:2409.02104},
  year={2024}
}

@article{jeong2024track4gen,
  title={Track4Gen: Teaching Video Diffusion Models to Track Points Improves Video Generation},
  author={Jeong, Hyeonho and Huang, Chun-Hao Paul and Ye, Jong Chul and Mitra, Niloy and Ceylan, Duygu},
  journal={arXiv preprint arXiv:2412.06016},
  year={2024}
}

@misc{wen2024anypointtrajectorymodelingpolicy,
      title={Any-point Trajectory Modeling for Policy Learning}, 
      author={Chuan Wen and Xingyu Lin and John So and Kai Chen and Qi Dou and Yang Gao and Pieter Abbeel},
      year={2024},
      eprint={2401.00025},
      archivePrefix={arXiv},
      primaryClass={cs.RO},
      url={https://arxiv.org/abs/2401.00025}, 
}

@InProceedings{shrivastava2024gmrw,
      title     = {Self-Supervised Any-Point Tracking by Contrastive Random Walks},
      author    = {Shrivastava, Ayush and Owens, Andrew},
      journal   = {European Conference on Computer Vision (ECCV)},
      year      = {2024},
      url       = {https://arxiv.org/abs/2409.16288},
}

@article{badki2025l4p,
  author    = {Badki, Abhishek and Su, Hang and Wen, Bowen and Gallo, Orazio},
  title     = {{L4P}: {L}ow-Level {4D} Vision Perception Unified},
  journal   = arxiv,
  year      = {2025},
}

@inproceedings{wang2024vggsfm,
  title={VGGSfM: Visual Geometry Grounded Deep Structure From Motion},
  author={Wang, Jianyuan and Karaev, Nikita and Rupprecht, Christian and Novotny, David},
  booktitle={Proceedings of the IEEE/CVF Conference on Computer Vision and Pattern Recognition},
  pages={21686--21697},
  year={2024}
}

@inproceedings{kastenfast,
  title={Fast Encoder-Based 3D from Casual Videos via Point Track Processing},
  author={Kasten, Yoni and Lu, Wuyue and Maron, Haggai},
  booktitle={The Thirty-eighth Annual Conference on Neural Information Processing Systems}
}

@inproceedings{rockwell2025dynpose,
   title     = {Dynamic Camera Poses and Where to Find Them},
   author    = {Rockwell, Chris and Tung, Joseph and Lin, Tsung-Yi and Liu, Ming-Yu and Fouhey, David F. and Lin, Chen-Hsuan},
   booktitle = {CVPR},
   journal   = {arXiv preprint arXiv:2504.17788},
   year      = {2025}
 }

@article{st4rtrack2025,
  title={St4RTrack: Simultaneous 4D Reconstruction and Tracking in the World},
  author={Feng, Haiwen and Zhang, Junyi and Wang, Qianqian and Ye, Yufei and Yu, Pengcheng and Black, Michael J. and Darrell, Trevor and Kanazawa, Angjoo},
  journal={arXiv preprint arxiv:2504.13152},
  year={2025}
}

@misc{huang2025segmentmotionvideos,
      title={Segment Any Motion in Videos}, 
      author={Nan Huang and Wenzhao Zheng and Chenfeng Xu and Kurt Keutzer and Shanghang Zhang and Angjoo Kanazawa and Qianqian Wang},
      year={2025},
      eprint={2503.22268},
      archivePrefix={arXiv},
      primaryClass={cs.CV},
      url={https://arxiv.org/abs/2503.22268}, 
}

@inproceedings{karazija24learning, title={Learning segmentation from point trajectories}, author={Karazija, Laurynas and Laina, Iro and Rupprecht, Christian and Vedaldi, Andrea}, booktitle={The Thirty-eighth Annual Conference on Neural Information Processing Systems}, year={2024} }

@article{cheng2021mask2former,
    title={Masked-attention Mask Transformer for Universal Image Segmentation},
    author={Bowen Cheng and Ishan Misra and Alexander G. Schwing and Alexander Kirillov and Rohit Girdhar},
    journal={arXiv},
    year={2021}
}

@article{peize2021dance,
  title   =  {DanceTrack: Multi-Object Tracking in Uniform Appearance and Diverse Motion},
  author  =  {Peize Sun and Jinkun Cao and Yi Jiang and Zehuan Yuan and Song Bai and Kris Kitani and Ping Luo},
  journal =  {arXiv preprint arXiv:2111.14690},
  year    =  {2021}
}

@inproceedings{lemoing2024dense,
  title = {Dense Optical Tracking: Connecting the Dots},
  author = {Le Moing, Guillaume and Ponce, Jean and Schmid, Cordelia},
  year = {2024},
  booktitle = {CVPR}
}

@article{sohn2020fixmatch,
  title={Fixmatch: Simplifying semi-supervised learning with consistency and confidence},
  author={Sohn, Kihyuk and Berthelot, David and Carlini, Nicholas and Zhang, Zizhao and Zhang, Han and Raffel, Colin A and Cubuk, Ekin Dogus and Kurakin, Alexey and Li, Chun-Liang},
  journal={Advances in neural information processing systems},
  volume={33},
  pages={596--608},
  year={2020}
}

@inproceedings{li2024blinkvision,
  title={Blinkvision: A benchmark for optical flow, scene flow and point tracking estimation using rgb frames and events},
  author={Li, Yijin and Shen, Yichen and Huang, Zhaoyang and Chen, Shuo and Bian, Weikang and Shi, Xiaoyu and Wang, Fu-Yun and Sun, Keqiang and Bao, Hujun and Cui, Zhaopeng and others},
  booktitle={European Conference on Computer Vision},
  pages={19--36},
  year={2024},
  organization={Springer}
}

@inproceedings{zholus2025tapnext,
  title={Tapnext: Tracking any point (tap) as next token prediction},
  author={Zholus, Artem and Doersch, Carl and Yang, Yi and Koppula, Skanda and Patraucean, Viorica and He, Xu Owen and Rocco, Ignacio and Sajjadi, Mehdi SM and Chandar, Sarath and Goroshin, Ross},
  booktitle={Proceedings of the IEEE/CVF International Conference on Computer Vision},
  pages={9693--9703},
  year={2025}
}
}

\newpage

\end{document}